\RequirePackage{fix-cm}
\documentclass[twocolumn]{svjour3}          
\smartqed  
\usepackage{graphicx}
\usepackage{natbib}
\usepackage{multirow}
\usepackage{amsmath}
\usepackage{amssymb}
\usepackage{bbm}
\usepackage{bm}
\usepackage{booktabs}
\usepackage{array}
\usepackage{blindtext}
\usepackage{comment}

\usepackage{pifont}
\usepackage[dvipsnames]{xcolor}
\usepackage{color, colortbl}
\definecolor{citecolor}{HTML}{0071bc}
\definecolor{tabhighlight}{HTML}{e5e5e5}
\usepackage[colorlinks,citecolor=citecolor]{hyperref}

\usepackage{wrapfig,caption,subcaption}
\usepackage{tabulary,xspace,makecell}
\usepackage{dsfont,fixmath,mathtools,nicefrac}

\def\vb{{\bm{b}}}
\def\vc{{\bm{c}}}

\def\ve{{\bm{e}}}

\def\vp{{\bm{p}}}
\def\vq{{\bm{q}}}

\def\vt{{\bm{t}}}

\def\vv{{\bm{v}}}
\def\vw{{\bm{w}}}
\def\vx{{\bm{x}}}

\def\vz{{\bm{z}}}

\newcommand{\rowNumber}[1]{\textcolor{Cerulean}{#1}}

\providecommand{\eg}{\textit{e.g.}\@\xspace}
\providecommand{\ie}{\textit{i.e.}\@\xspace}
\newcommand{\cmark}{\ding{51}}
\newcommand{\xmark}{\ding{55}}
\newcommand{\snowflake}{\includegraphics[width=10px]{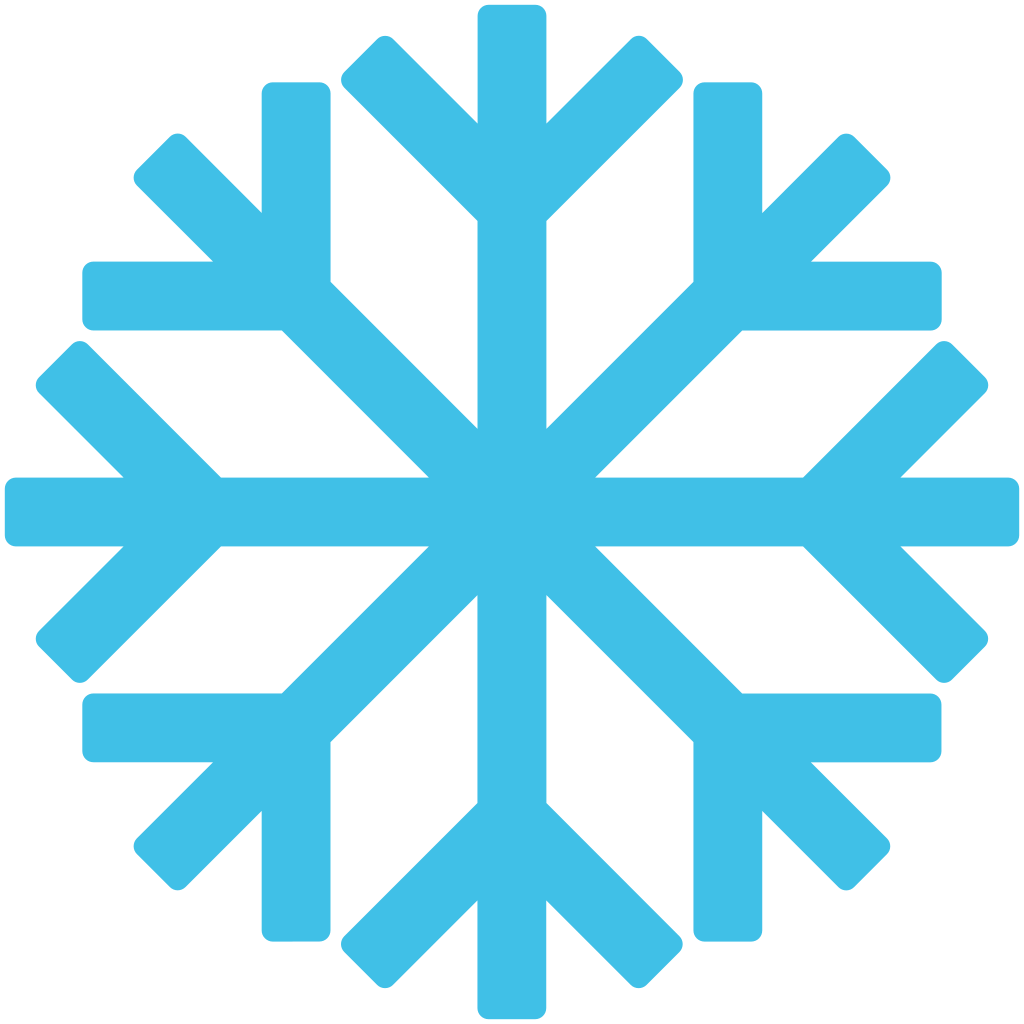}}

\newcommand{\methodname}{ContextDET\xspace}
\newcommand{\benchmarkname}{CODE\xspace}

\definecolor{brinkpink}{rgb}{0.98, 0.38, 0.5}

\makeatletter
\renewcommand\paragraph{
  \@startsection{paragraph} 
  {4} 
  {\z@} 
  {.5em \@plus1ex \@minus.2ex} 
  {-.5em} 
  {\normalfont\normalsize\bfseries} 
}
\makeatother
\begin{document}
\sloppy

\title{Contextual Object Detection with Multimodal Large Language Models}


\author{Yuhang Zang \and
        Wei Li \and
        Jun Han \and
        Kaiyang Zhou \and
        Chen Change Loy
}

\institute{Yuhang Zang \at
              S-Lab, Nanyang Technological University, Singapore \\
              \email{zang0012@ntu.edu.sg}
           \and
           Wei Li \at
              S-Lab, Nanyang Technological University, Singapore \\
              \email{wei.l@ntu.edu.sg}
           \and
           Jun Han \at
              S-Lab, Nanyang Technological University, Singapore \\
              \email{jun.han@ntu.edu.sg}
           \and
           Kaiyang Zhou \at
              Hong Kong Baptist University, Hong Kong \\
              \email{kyzhou@hkbu.edu.hk}
           \and
           Chen Change Loy (corresponding author) \at
              S-Lab, Nanyang Technological University, Singapore \\
              \email{ccloy@ntu.edu.sg}
}

\date{Received: date / Accepted: date}
\maketitle
\begin{abstract}
Recent Multimodal Large Language Models (MLLMs) are remarkable in vision-language tasks, such as image captioning and question answering, but lack the essential perception ability, \ie, object detection. In this work, we address this limitation by introducing a novel research problem of \textit{contextual object detection}---understanding visible objects within different human-AI interactive contexts. Three representative scenarios are investigated, including the language cloze test, visual captioning, and question answering. Moreover, we present ContextDET, a unified multimodal model that is capable of end-to-end differentiable modeling of visual-language contexts, so as to locate, identify, and associate visual objects with language inputs for human-AI interaction. 
Our ContextDET involves three key submodels: (i) a visual encoder for extracting visual representations, (ii) a pre-trained LLM for multimodal context decoding, and (iii) a visual decoder for predicting bounding boxes given contextual object words. The new \textit{generate-then-detect} framework enables us to detect object words within human vocabulary. Extensive experiments show the advantages of ContextDET on our proposed CODE benchmark, open-vocabulary detection, and referring image segmentation.
\end{abstract}

\begin{figure*}[!t]
    \centering
    \includegraphics[width=.98\textwidth]{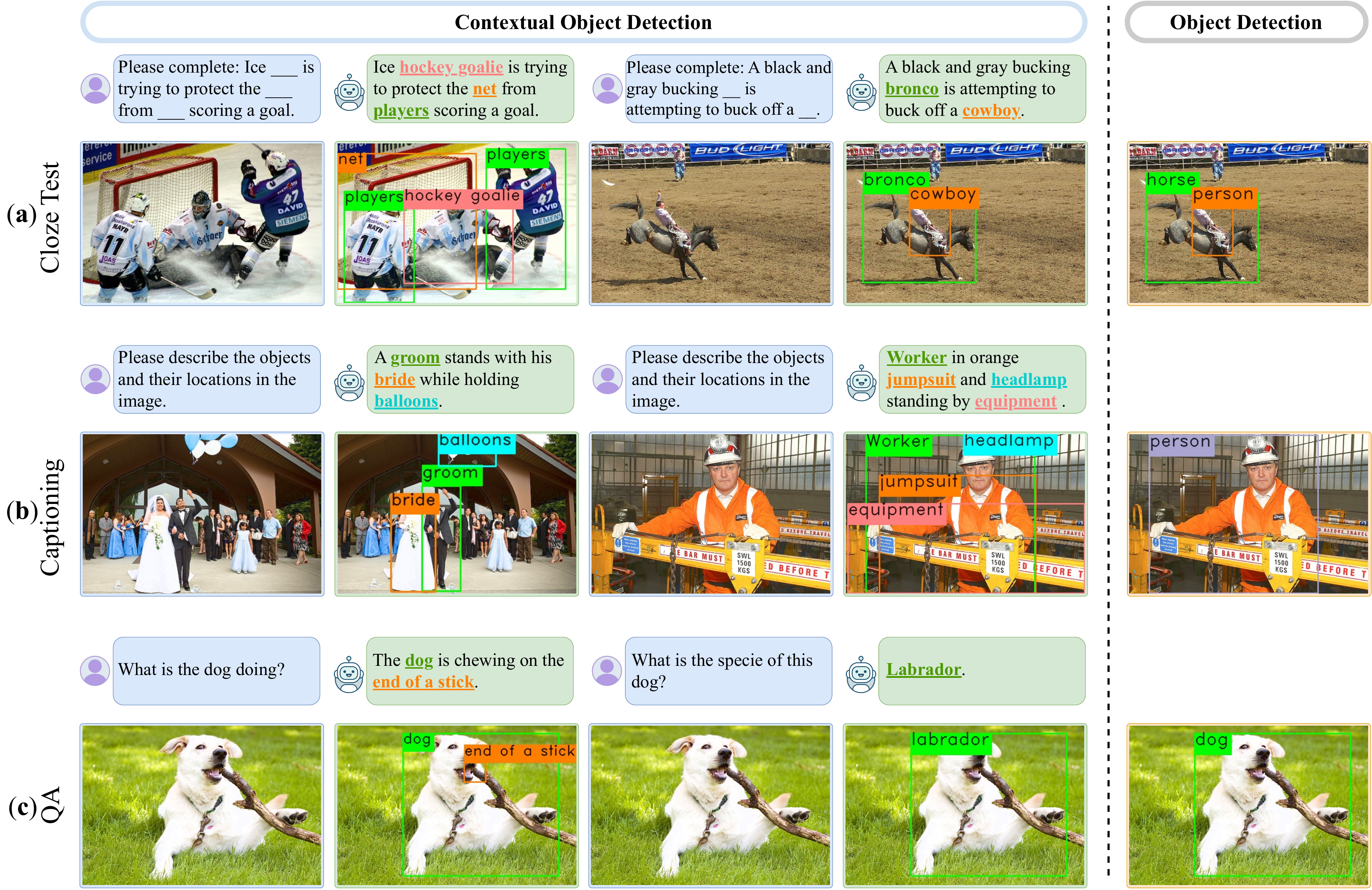}
    \caption{\small
    We present a new \textbf{contextual object detection} task include \textbf{(a)} look at the image and complete the masked object names and locations; \textbf{(b)} predict the caption and the boxes of objects existing in the caption; \textbf{(c)} answer a question about the names and locations of objects.
    Unlike the traditional object detection task that typically focuses on detecting a limited set of pre-defined object classes such as `\textcolor{Blue!40}{person}', our task requires predicting more specific names (\eg, `\textcolor{brinkpink}{hockey goalie}', `\textcolor{ForestGreen!90}{groom}', or `\textcolor{orange!90}{bride}') based on contextual understanding.
    }
    \label{fig:code_teaser}
\end{figure*}

\section{Introduction}

``\emph{For me context is the key - from that comes the understanding of everything.}'' --~Kenneth Noland

One indispensable cornerstone of computer vision---object detection---is understanding visible objects within scenes, which empowers many applications, such as robotics, autonomous driving, and AR/VR systems.  Recently, Multi-modal Language Models (MLLMs) trained with internet-scale visual-language data, including Flamingo~\citep{Alayrac2022FlamingoAV}, PaLM-E~\citep{driess2023palme}, and the superb OpenAI's GPT-4~\citep{openai2023gpt4}, have shown a revolutionary ability to allow humans to interact with AI models for various vision-language tasks, \eg, image captioning and question answering. 
Such an interactive human-AI circumstance requires modeling \textit{contextual} information, \ie, relationships among visual objects, human words, phrases, and even dialogues. Therefore, it is desirable to advance MLLMs with the capability of locating, identifying, and associating visual objects with language inputs for human-AI interaction. 

In this paper, we study a new research problem---contextual object detection---that is understanding visible objects within human-AI interactive contexts. In comparison with existing standard object detection, we consider four comprehensive objectives for such a new setting: (i) \textbf{\textit{capacity}}: being able to handle a human language vocabulary; (ii) \textbf{\textit{description}}: describing visual inputs from users with informative natural language statements; (iii) \textbf{\textit{perception}}: locating and associating visual objects with language queries; (iv) \textbf{\textit{understanding}}: complementing proper words with language hints. To cover these four objectives, we incorporate three representative tasks: language cloze test, visual captioning, and question answering, with object detection for MLLMs (see Fig.~\ref{fig:code_teaser}).

While significant progress has been made in developing more accurate and faster object detection algorithms, it is impossible to directly integrate existing deep object detectors with MLLMs for contextual object detection, due to the following reasons.  First, standard deep detectors, such as Mask-RCNN~\citep{he2017mask} and DETR~\citep{carion2020end}, are trained with close-set classifiers and cannot generalize well in real-world scenarios, where object categories or classes are not pre-defined or limited to a closed set. Despite the very recent development of open-vocabulary object detection~\citep{gu2021open,zhou2022detecting,zang2022open,rasheed2022bridging} that builds on state-of-the-art vision-language models (e.g., CLIP~\citep{radford2021learning} and ALIGN~\citep{jia2021scaling}) can improve the zero-shot transfer ability for novel classes, they are constrained by the scale of pre-defined novel categories, making them incapable of detecting objects for a human language vocabulary. While some papers \citep{dai2023exploring} explore using LLMs to improve binary OOD classification, recognizing novel class names in human language vocabulary has not been addressed. For example, these open-vocabulary detectors fail to handle out-of-distributed categories in Fig.~\ref{fig:code_teaser}, such as hockey goalie, groom, and cowboy. Second, the inherent \textit{locate-then-classify} paradigm of existing deep detection models is unsuitable for contextual object detection. In generic human-AI interactive scenarios, both natural objects in visual scenes and human words in language inputs have various meanings in different contexts. In Fig.~\ref{fig:code_teaser} (a) and (b), the universal `person' category will manifest as `goalie', `player', `cowboy', `groom', `bride', and `worker' within distinct visual contexts. Also, as language contexts shift, the word `labrador' supplants the representation of `dog' (Fig.~\ref{fig:code_teaser} (c)). 
Consequently, an innovative detection approach is required to cater to considerably varied and changing contextual object detection.

\begin{table*}[t]
    \centering
    \small
    \setlength{\abovecaptionskip}{0mm}
    \caption{\small
    Comparison of our proposed three contextual object detection settings with previous related tasks.
    }
    \label{tab:related_work}
    \setlength\tabcolsep{4.5pt}
    \resizebox{.9\textwidth}{!}{
    \begin{tabular}{l ccc}
    \toprule
    Tasks  & Language Input & Output(s) & Remark \\
    \midrule
      Object Detection &  \xmark & box, class label & pre-defined class labels \\
      \cmidrule(lr){1-1} \cmidrule(lr){2-2} \cmidrule(lr){3-3} \cmidrule(lr){4-4}
      Open-Vocabulary Object Detection & (optional) class names for CLIP & box, class label & pre-defined class labels \\
      \cmidrule(lr){1-1} \cmidrule(lr){2-2} \cmidrule(lr){3-3} \cmidrule(lr){4-4}
      Referring Expression Comprehension & complete referring expression & box that expression refers to & / \\
      \cmidrule(lr){1-1} \cmidrule(lr){2-2} \cmidrule(lr){3-3} \cmidrule(lr){4-4}
      \cmidrule(lr){1-1} \cmidrule(lr){2-2} \cmidrule(lr){3-3} \cmidrule(lr){4-4}
      \rowcolor{violet!10}
      \textbf{\quad Contextual} & \textcolor{orange}{\textbf{incomplete}} expression& \{box, \textcolor{orange}{\textbf{name}}\} & \textcolor{orange}{\textbf{name}}\ could be \\
      \rowcolor{violet!10}
      \textbf{Cloze Test} (ours) & object names are masked & to complete the mask & most valid English word
      \\
      \midrule
      Image Captioning & \xmark & language caption \\
      \rowcolor{violet!10}
      \textbf{Contextual Captioning} (ours) & \xmark & language caption, \textcolor{orange}{\textbf{box}} & \\
      \midrule
      Visual Question Answering & language question & language answer \\
      \rowcolor{violet!10}
      \textbf{Contextual QA} (ours) & language question & language answer, \textcolor{orange}{\textbf{box}} & \\
    \bottomrule
    \end{tabular}}
\end{table*}

To address the above challenges, in this work, we present \methodname, a novel \textit{generate-then-detect} framework, specialized for contextual object detection. Specifically, it is an end-to-end model that consists of three key modules. First, a visual encoder extracts high-level image representations for given images and produces both local and full visual tokens for further contextual modeling. Second, to effectively model multimodal contexts, we employ a pre-trained LLM to perform text generation, with conditioned inputs of both local visual tokens and task-related language tokens as the multimodal prefix. Third, taking the LLM tokens as prior knowledge for visual detection, we introduce a visual decoder that consists of multiple cross-attention layers, within which we compute conditional object queries from contextual LLM tokens, and keys and values from full visual tokens, to predict the corresponding matching scores and bounding boxes. This allows us to detect contextual object words for a human vocabulary.

\noindent \textbf{Contributions.} In summary, our contributions are the following: (\textbf{i}) We investigate contextual object detection---a new direction for visual object detection that improves MLLMs with a greater ability for human-AI interaction. (\textbf{ii}) To open this area to empirical study, we present a new benchmark CODE with 10,346 unique object words to facilitate research on contextual object detection. (\textbf{iii}) We propose a novel \textit{generate-then-detect} framework, ContextDET, dedicated to contextual object detection. (\textbf{iv}) We demonstrate the advantages of our ContextDET not only on the CODE benchmark but also on open-vocabulary detection and referring image segmentation tasks. We hope our work can motivate future research in contextual object detection that benefits human-AI interaction.

\section{Related Work}
\noindent \textbf{Multimodal Large Language Models (MLLMs).} Large Language Models (LLMs) have been developed to comprehend and generate textual language, showcasing remarkable performance across a wide range of Natural Language Processing (NLP) tasks. Notable examples of LLMs include OpenAI's GPT series \citep{radford2018improving, radford2019language, brown2020language, chatgpt2022, openai2023gpt4}, Google's T5 \citep{raffel2020exploring} and PaLM \cite{chowdhery2022palm}, as well as Meta's OPT \citep{zhang2022opt} and LLaMA \citep{touvron2023llama}. More recently, there have been advancements in the field of MLLMs \citep{mokady2021clipcap, tsimpoukelli2021multimodal, chen2022visualgpt, koh2023grounding, li2023blip2, huang2023language, driess2023palme, openai2023gpt4}, exemplified by the GPT-4 model \citep{openai2023gpt4}, which have expanded the capabilities of LLMs to comprehend both language and visual inputs. MLLMs have demonstrated impressive proficiency in a range of vision-language tasks, including image captioning and visual question answering. However, existing MLLMs are limited to generating textual outputs. In contrast, our \methodname, built upon MLLMs, extends support to contextual object detection, providing bounding box outputs. Further comparisons are discussed in Sec.~\ref{sec:qualitative_results}.

\noindent \textbf{Prompting LLMs with Vision Experts}. Several recent papers~\citep{shen2023hugginggpt, wu2023visual, yang2023mm} have proposed systems that leverage the textual output generated by LLMs, such as ChatGPT~\citep{chatgpt2022}, as prompts to manipulate external vision expert models for various vision-related tasks. In the context of object detection, these vision expert models include DETR~\citep{carion2020end}, Grounding DINO~\citep{liu2023grounding}, SAM~\citep{kirillov2023segment}, and other algorithms integrated into the HuggingFace community~\citep{huggingface}. However, due to the frozen parameters of both LLMs and expert models, the knowledge and representations from LLMs cannot be shared, potentially leading to sub-optimal performance. In contrast to these prompting-based methods, our \methodname employs an end-to-end training pipeline. We utilize the latent features extracted from MLLMs as conditional inputs for a visual decoder, enabling the prediction of bounding boxes.

\noindent \textbf{Object Detection with Contextual Understanding}. The term ``context'' commonly refers to the neighboring pixels or surrounding regions within images and has been extensively explored in previous studies to enhance object detection algorithms~\citep{divvala2009empirical, mottaghi2014role, shrivastava2016contextual, chen2018context}. In this paper, the concept of contextual information encompasses multimodal patterns and relationships, involving both visual images and textual words. Our \methodname leverages the robust contextual understanding capability of MLLMs and applies it to the downstream object detection task. Additionally, we propose the adoption of new evaluation tasks, such as the cloze test, to more effectively assess the contextual understanding ability.

\noindent \textbf{Object Detection on Novel Classes}. Despite significant advancements in deep learning techniques~\citep{ren2015faster, liu2016ssd, law2018cornernet, tian2019fcos, carion2020end, chen2021pix2seq, liu2021swin, zhang2022dino, zhudeformable,wang2022internimage}, object detection remains a challenging task in real-world scenarios, particularly in the case of zero-shot object detection~\citep{bansal2018zero}. Zero-shot object detection requires models trained on \emph{base} classes to detect \emph{novel} classes that were not encountered during training. A recent variant of zero-shot detection, known as Open-Vocabulary Object Detection, allows for the utilization of additional image-text pairs~\citep{zareian2021open}, garnering significant attention from the research community. In this context, recent vision and language pre-trained models~\citep{radford2021learning, zhong2022regionclip, li2022grounded, zhang2022glipv2}, such as CLIP, have been widely employed for open-vocabulary object detection~\citep{gu2021open, zhou2022detecting, du2022learning,zang2022open, rasheed2022bridging,kuo2022f,wu2023betrayed,wu2023aligning,wu2023cora}. Instead of relying solely on CLIP, our \methodname demonstrates that MLLMs can also be applied effectively to the open-vocabulary setting. With the assistance of MLLMs, \methodname is not constrained by pre-defined \emph{base} or \emph{novel} classes. Notably, the object names predicted by \methodname can be generated as the most contextually valid English words by the MLLMs.

\noindent \textbf{Visual Grounding}. Visual grounding tasks, such as referring expression comprehension~\citep{karpathy2015deep}, involve combining object detection with language understanding abilities. In these tasks, a language query is provided to describe a specific object, and models are tasked with predicting the position of the referred object. State-of-the-art algorithms~\citep{yang2022lavt,wang2022cris} commonly employ Transformer-based cross-modal structures or multimodal pre-training~\citep{kamath2021mdetr}. Our proposed contextual object detection task presents even greater challenges compared to visual grounding. For example, in our cloze test, the language query is incomplete, and the object names are masked. Models are required to infer both the missing object name words and their positions based on contextual information. Furthermore, in our contextual captioning setting, no language query is given. Additionally, in our contextual QA setting, the objects are described using human language in an \emph{interactive} environment.

\noindent \textbf{Image Captioning.} Image captioning focuses on generating descriptive sentences to understand given images. Typically, image captioning models first encode the input image as feature embeddings using pre-trained classification~\cite{chen2017sca}, object detection~\cite{anderson2018bottom} or vision language models~\cite{mokady2021clipcap}. Subsequently, submodules like LSTMs~\cite{hochreiter1997long} or Transformers~\cite{vaswani2017attention} are employed to decode feature embeddings into predicted sentences. In contrast, our \emph{contextual captioning} task extends beyond language outputs by requiring the model to predict the locations of the bounding boxes containing the objects mentioned in the generated captions.

\noindent \textbf{Visual Question Answering (VQA).} Visual question answering tasks involve answering questions related to given images~\cite{antol2015vqa,goyal2017making}. In traditional VQA, model inputs and outputs are comprised of question-answer pairs in natural language. However, in our \emph{contextual QA} task, questions are specifically focused on inquiring about object names and locations, while corresponding answers are expected to include the corresponding referring bounding boxes.

\section{Approach}\label{sec:approach}

\begin{figure*}[t]
    \centering
    \includegraphics[width=.95\textwidth]{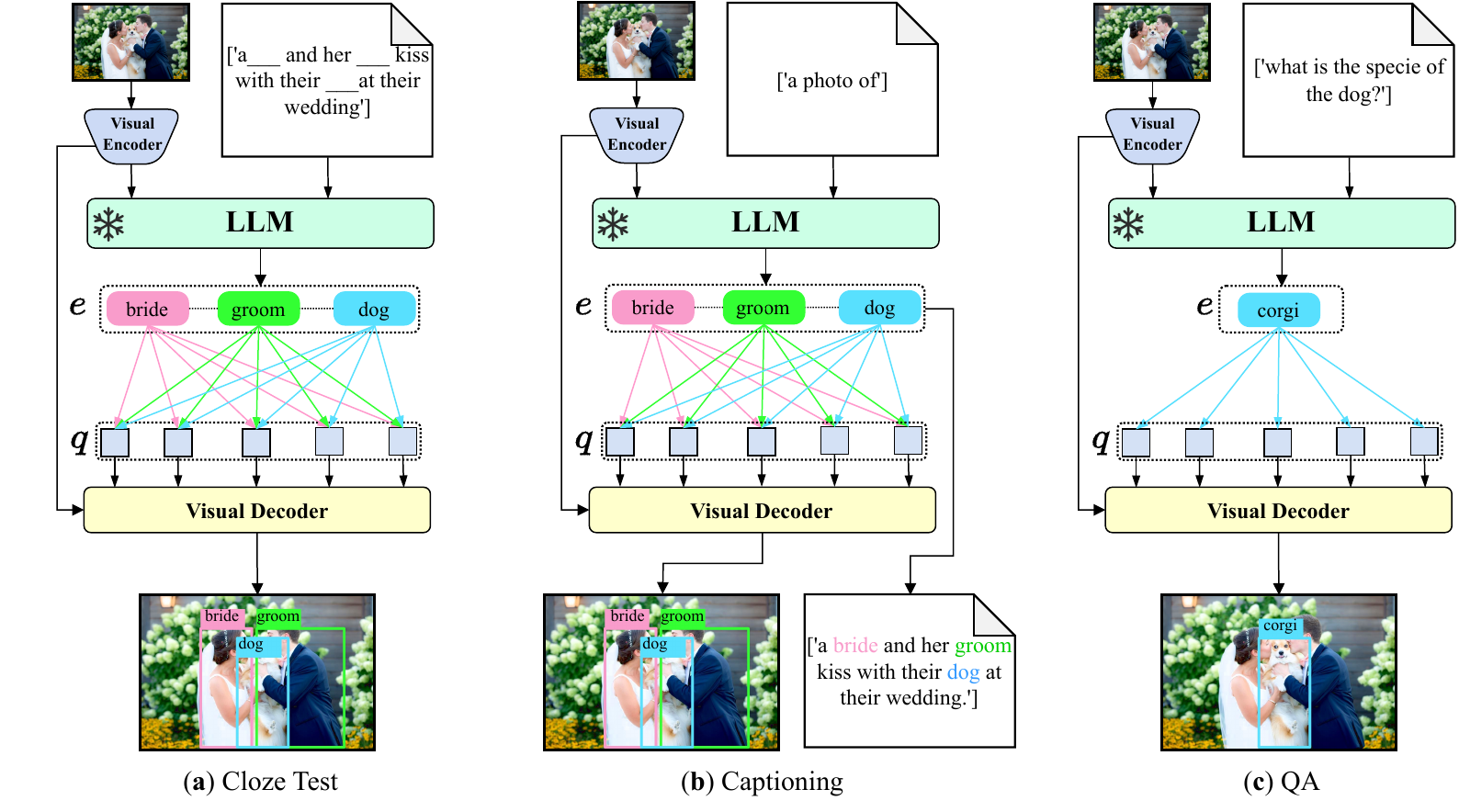}
    \caption{ \small
    Our \methodname is a unified end-to-end framework, being capable of taking different language token inputs for different tasks, including \textbf{(a)} cloze test \textbf{(b)} captioning and \textbf{(c)} question answering.
    \snowflake: frozen.
    The symbol $\ve$ indicates latent embeddings of LLM (Sec.~\ref{sec:LLM_with_multi_modal}), and the symbol $\vq$ denotes object queries of the visual decoder (Sec.~
\ref{sec:visual_decoder}).
    }
    \label{fig:contextdet_pipeline}
    \vspace{-6pt}
\end{figure*}

This section describes our contextual object detection framework, \methodname, which accepts images interleaved with human text as inputs and produces free-form text and corresponding bounding boxes as outputs. As illustrated in Fig.~\ref{fig:contextdet_pipeline}, our \methodname is end-to-end and consists of three key architectural components: (1) a visual encoder that extracts high-level image representations and computes visual tokens, (2) a pre-trained LLM that decodes multimodal contextual tokens with a task-related multimodal prefix, and (3) a visual decoder that predicts matching scores and bounding boxes for conditional queries linked to contextual object words.

\subsection{Visual Encoder}\label{sec:visual_encoder}

Given an image input $\vx \in \mathbb{R}^{3 \times H \times W}$, we use a vision backbone parameterized by $\phi$ to extract image-level spatial features $\vv = f_{\phi}(\vx) \in \mathbb{R}^{d \times h \times w}$, where $d$ denotes the feature dimension. The vision backbone $\phi$ is pre-trained and frozen, which can be selected from various options, including ResNet~\citep{he2016deep}, Vision Transformer~(ViT)~\citep{dosovitskiy2021image}, or Swin Transformer~\citep{liu2021swin}.
Subsequently, the image-level features $\vv$ are transformed into two distinct representations.

\noindent \textbf{Local Visual Tokens.} We first divide the 2D spatial grid of features as $p$ local bins and apply adaptive average pooling for each bin, followed by a linear projection then flattened to 1D: $\vz = \texttt{Linear}({\texttt{AvgPool}(\vv)})$. As a result, fixed-sized visual tokens $\vz \in \mathbb{R}^{d_{1} \times p}$ are obtained and fed to the LLM (Sec.~\ref{sec:LLM_with_multi_modal}), Here, $d_{1}$ represents the input dimension of LLM.  

\noindent \textbf{Full Visual Tokens.} We flatten the 2D spatial features $\vv$ as 1D sequence with $m = h \times w$ tokens and leverage six Transformer layers $\psi$ to compute the encoded full visual tokens: $\vc = f_{\psi}(\vv) \in \mathbb{R}^{d_{2} \times m}$, which will serve as inputs for the visual decoder (Sec.~\ref{sec:visual_decoder}).

\subsection{Multimodal Context Modeling with LLM} \label{sec:LLM_with_multi_modal}
Motivated by the finding that LLMs are strong context generators~\citep{yu2023generate} for solving various knowledge-intensive tasks, it is thus appealing to model multimodal contexts with LLMs. We consider performing text generation with the LLM, conditioned on both the visual representations produced by the visual encoder described in Sec.~\ref{sec:visual_encoder} and task-oriented human languages.

\noindent \textbf{Multimodal Tokens.} Given the visual context of input images, we generate language contexts that describe the visual information or complement missing words. Specifically, the inputs to the LLM consist of (1) the local visual tokens $\vz \in \mathbb{R}^{d_{1} \times p}$, and (2) a series of language tokens $\vt_{1:l} = \{\vt_{1},\ldots,\vt_{l}\} \in \mathbb{R}^{d_{1} \times l}$, where the symbol $l$ is the sequence length of the language tokens.  The language tokens $\vt_{1:l}$
have different forms for different contextual object detection settings. For the cloze test, the language tokens are tokenized  sentences with masked names, \eg, `\texttt{a [MASK] and her [MASK] kiss with their [MASK] at their wedding}'.
For the visual captioning, the language tokens are tokenized text prompts---`\texttt{a photo of}'---to describe the image. For the question answering, the language tokens represent the tokenized sentences of questions, \eg, `\texttt{Question: what is the specie of the dog? Answer:}'.

\noindent \textbf{Multimodal Prefixed LLM Decoding.} A pre-trained LLM $\theta$ can be conditioned on a prefix 
$\vw_{1:n}$ 
that contains multimodal tokens to generate text in an autoregressive way:
\begin{align}
p(\vw_{n+1:L}|\vw_{1:n}) = \prod_{i=n+1}^{L}p_{\theta}(\vw_{i+1}|\vw_{1:i}).
\end{align}
Here, the prefix $\vw_{1:n}=[\vz,\vt_{1:l}] \in \mathbb{R}^{d_{1} \times (p+l)}$ is obtained via concatenating the local visual tokens $\vz$ with a sequence of language tokens $\vt_{1:l}$. 
Specifically, the LLM consists of multiple Transformer layers (\texttt{TransLayers}) and a final Feed Forward Network (\texttt{FFN}). 
To generate new tokens, the LLM first predicts the latent embedding $\ve_{n+1}$ for the new $n+1$-th token:
\begin{align} \label{eq:llm_latent}
\ve_{n+1} = \texttt{TransLayers}( \vw_{1:n} ),
\end{align}
which contains decoded multimodal contextual information.
Then, the \texttt{FFN} computes the probability distribution $p(\vw_{n+1})$ based on the latent embedding $\ve_{n+1}$:
\begin{align}
p(\vw_{n+1}) = \texttt{Softmax}(\texttt{FFN}(\ve_{n+1})),
\end{align}
where the tokens $\vw_{n+1}$ are elements of a vocabulary $\mathcal{W}$ that corresponding to human words in natural language. Such autoregressive generation ends when the generated language token $\vw_{L}$ hits the \texttt{[EOS]} token, \ie, the ending of sentences. 

\subsection{Visual Decoder} \label{sec:visual_decoder}
In order to associate object words with corresponding visual objects in given images, we propose a novel \textit{generate-then-detect} pipeline for contextual object detection. Unlike the common \textit{detect-then-classify} pipeline in standard object detectors (\eg, Mask R-CNN~\citep{he2017mask} and DETR~\citep{carion2020end}) that exhaustively locate and recognize all possible objects as pre-defined categories, we consider using the LLM tokens as prior knowledge for visual detection. This allows us to detect contextual object words, while not being limited to a close set of object classes.

\noindent \textbf{Contextual LLM Tokens as Conditional Object Queries.}
From both language prefix $\vt_{1:l}$ and generated tokens $\vw_{n+1:L}$ (Sec.~\ref{sec:LLM_with_multi_modal}), we predict the binary-classification probability of noun object words.
Then, we automatically select those language tokens related to object words (\eg, `bride', `groom', `dog') as contextual object tokens and 
take their latent embeddings as conditional inputs for the visual decoder.
To be specific, we use a binary label $\bar{\vw}$ to indicate whether a token belongs to an object word or not, which will be explored in Eq. (\ref{eq:loss}).
Then we set up $N$ learnable object queries $\vq$
as learnable positional embeddings in the visual decoder. For a contextual token, \eg, `bride', we obtain the conditional object queries that linked to `bride', by incorporating the corresponding latent embedding $\ve$ from the LLM with the object queries:
\begin{equation}
\bar{\vq} = \vq + \texttt{Linear}(\texttt{Repeat}(\ve)).
\end{equation}
Here, we repeat the latent embedding $\ve$ for `bride' $N$ times so as to align with the number of the object queries $\vq$. Also, a linear layer is employed for dimension projection.

\noindent \textbf{Conditional Multimodal Context Decoding.} To model cross-modal contextual relationships, we employ six Transformer cross-attention layers in the visual decoder, in which the keys and values are obtained from the full visual tokens $\vc$ extracted by the visual encoder (Sec.~\ref{sec:visual_encoder}) while the queries are derived from the conditional object queries $\bar{\vq}$ for computing cross-attention:
\begin{align}
\hat{\vq} = \texttt{CrossAttenLayers}(\vc, \bar{\vq}).
\end{align}
By doing so, the visual decoder learns to focus on specific areas of the visual context that are relevant to the conditional query for `bride'.

\noindent \textbf{Box and Matching Predictions for Contextual Words.} 
Finally, we compute the binary matching score and box prediction from the output latent embedding $\hat{\vq}$ using two $\texttt{FFN}$ prediction heads:
\begin{equation}
        \vp = \texttt{FFN}_{\text{cls}}(\hat{\vq}) \in \mathbb{R}^{N \times 2}, \vb = \texttt{FFN}_{\text{box}}(\hat{\vq}) \in \mathbb{R}^{N \times 4},
\end{equation}
where $\vp$ refers to the probability of being matched or not matched given the conditional object word, and $\vb$ indicates the predicted box coordinates.

\noindent \textbf{Conditional Matching for Label Assignment.} We introduce a conditional modification to the default optimal bipartite matching in DETR~\citep{kamath2021mdetr} that finds the best match between the set of $N$ predictions and the set of ground truth objects. 
In our approach, only the ground-truth bounding boxes that match the conditional object words are involved in the loss computation.
This conditional matching ensures that the model focuses solely on the objects described by the language queries.

\begin{figure*}[t]
    \centering
    \includegraphics[width=0.9\textwidth]{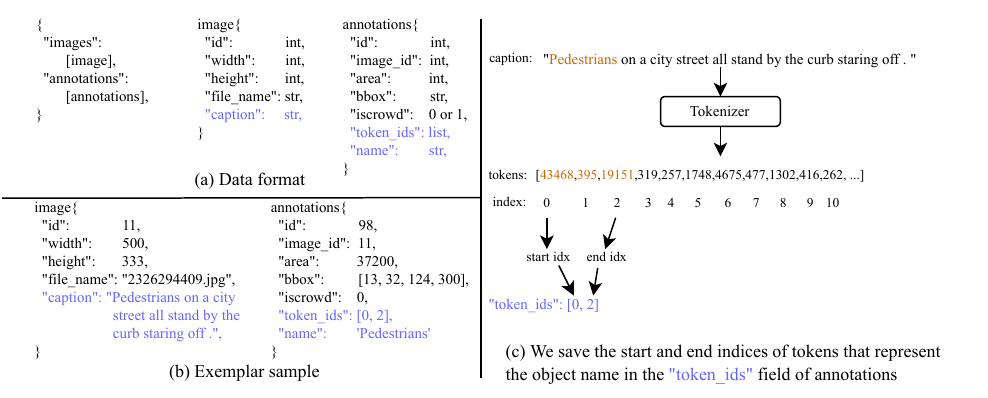}
    \caption{
    \small
    Our \benchmarkname benchmark follows the data format of the COCO dataset~\citep{lin2014microsoft}, with additional fields (\textcolor{blue!70}{blue} color) including the language caption, token ids, and object name. Token ids record the start and end position index of the object name existing in the language tokens.
    }
    \label{fig:contextdet_appendix_dataformat}
\end{figure*}

\begin{figure*}[t]
\begin{minipage}{0.32\textwidth}
\centering
\includegraphics[width=\linewidth]{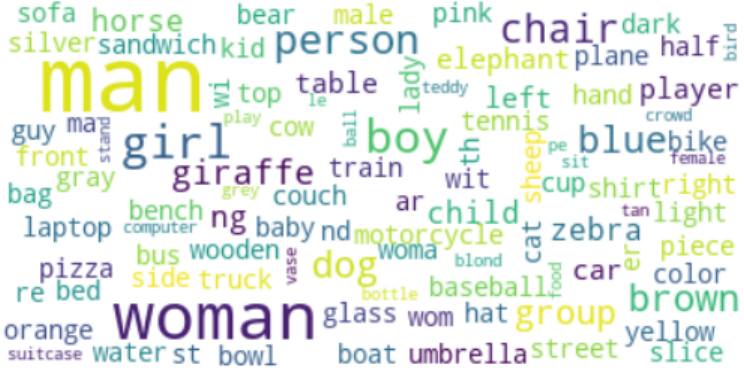}
\end{minipage}
\hfill
\begin{minipage}{0.32\textwidth}
\centering
\includegraphics[width=\linewidth]{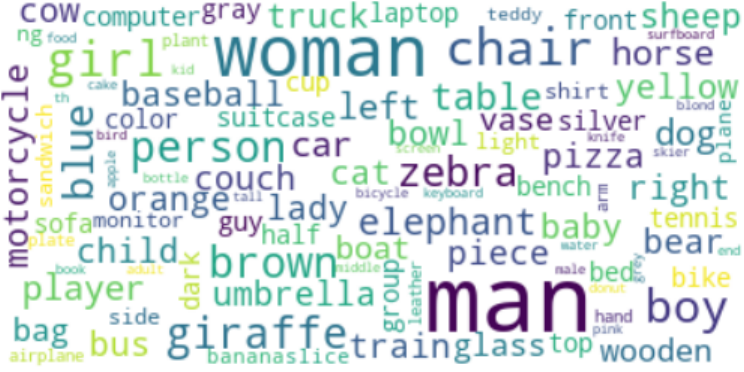}
\end{minipage}
\hfill
\begin{minipage}{0.32\textwidth}
\centering
\includegraphics[width=\linewidth]{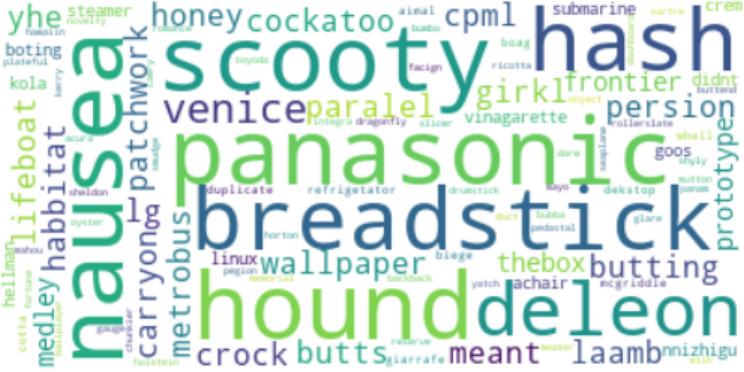}
\end{minipage}
\caption{
\small Word clouds of object words presented in the \benchmarkname train set (\textbf{left}) and test set (\textbf{middle}, \textbf{right}). The \textbf{middle} figure represents the visualization of high-frequency words in the test set, while the \textbf{right} figure showcases the visualization of low-frequency words.
}
\label{fig:appendix_word_cloud}
\end{figure*}

\subsection{Training Details}
\label{sec:training_details}

We use multi-scale deformable attention~\citep{zhudeformable} and IoU-based label assignment~\citep{ouyangzhang2022nms} to accelerate the convergence speed.
The vision encoder $\phi$ also supports the pre-trained weights from previous MLLM such as BLIP-2~\citep{li2023blip2}.

\noindent \textbf{Loss Function.}
In Sec.~\ref{sec:visual_decoder}, we use conditional matching to derive the label assignments, which include the ground-truth matching labels $\bm{\hat{\vp}}$ and the associated box coordinates $\bm{\hat{b}}$.
For our predicted language token $\vw$, we can straightforwardly get the annotated ground truth token $\hat{\vw}$, \eg, tokenized answers for the cloze test.
We can also obtain the annotated binary label $\bar{\vw}$ indicating whether a token belongs to an object word or not.
Based on the label assignment results, the overall loss function $\mathcal{L}$ is defined as:
\begin{equation}
\label{eq:loss}
\begin{aligned}
    \mathcal{L}&=\lambda_{\text{cls}} \mathcal{L}_{\text{cls}}\left(\vp, \hat{\vp} \right)+\lambda_{\text{box}}\mathcal{L}_{\text{box}}(\bm{b}, \hat{\bm{b}})\\ &+ \lambda_{\text{lm}}\mathcal{L}_{\text{lm}}(\bm{w}, \hat{\bm{w}})
+ \lambda_{\text{noun}}\mathcal{L}_{\text{noun}}(\vw, \bar{\vw})
\end{aligned}
\end{equation}
Here, the classification loss $\mathcal{L}_{\text{cls}}$ is a binary softmax classification loss of two classes: matched \textit{vs.} not matched.
The box-related loss $\mathcal{L}_{\text {box}}$ is either L1 loss or GIoU loss~\citep{rezatofighi2019generalized}.
The language modeling loss $\mathcal{L}_{\text{lm}}$ is softmax classification loss over the vocabulary size $\mathcal{W}$ of the LLM Tokenizer.
The noun loss $\mathcal{L}_{\text{noun}}$ is a binary classification loss that determines whether a token is an object word or not.
We set the loss weighting hyper-parameters $\lambda_{\text{cls}}=1$, $\lambda_{\text{box}}=5$, $\lambda_{\text{lm}}=1$, and $\lambda_{\text{noun}}=1$.

\begin{figure*}
    \centering
    \includegraphics[width=0.9\textwidth]{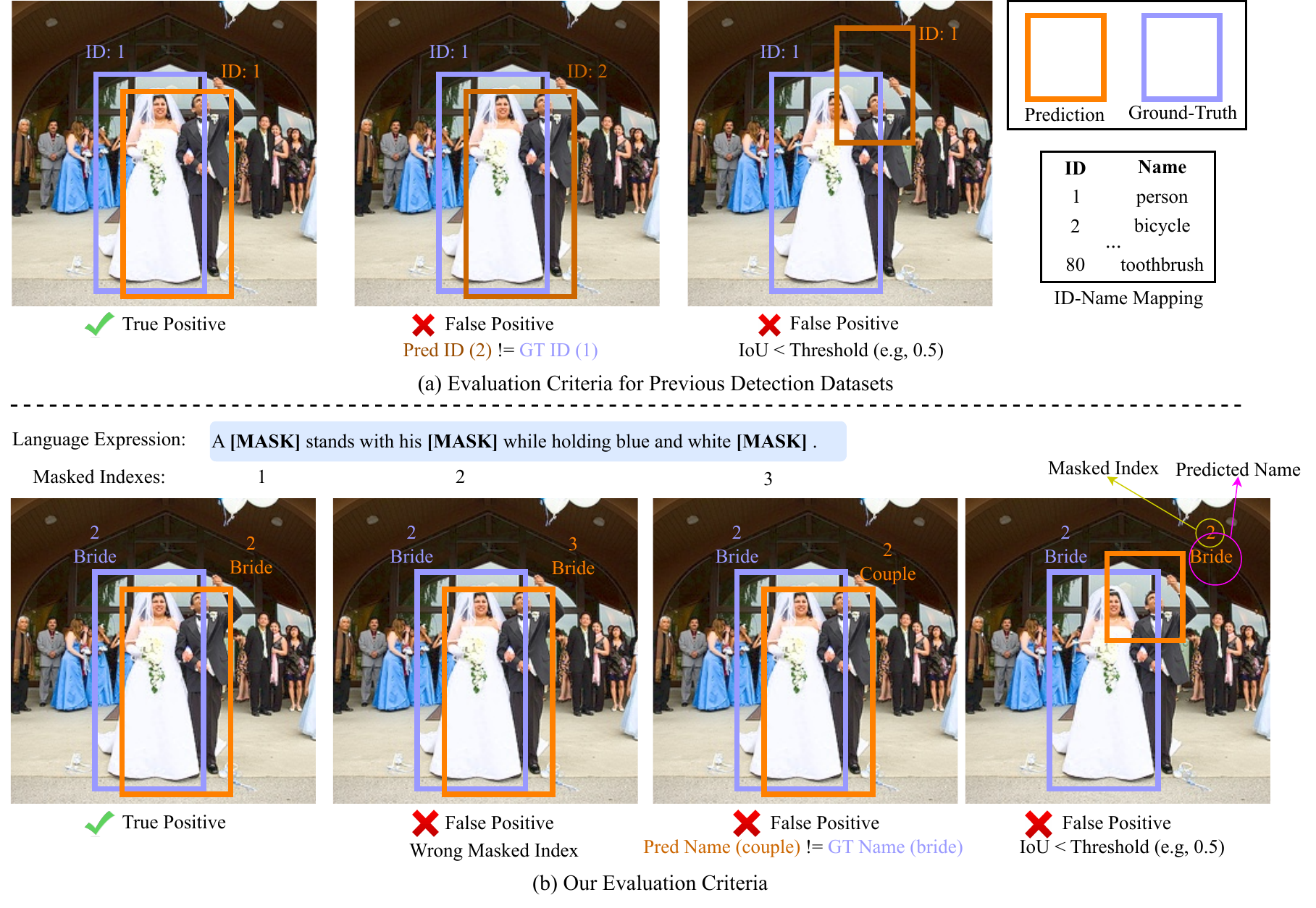}
    \caption{
    \small
    The comparison of (\textbf{a}) the evaluation criteria for the traditional object detection task, and (\textbf{b}) evaluation of our contextual cloze test.
    }
    \label{fig:contextdet_appendix_evaluation}
\end{figure*}

\section{CODE Benchmark} \label{appendix:dataset_details}

To facilitate research on contextual object detection, we construct a {C}ontextual {O}bject {DE}tection ({CODE}) dataset. Specifically, we collected images, bounding boxes and captions annotations from Flickr30k~\citep{young2014image} and Flickr30k Entities~\citep{plummer2015flickr30k}. We added annotations containing the position information of object names in the caption strings. These object names will be replaced with `\texttt{[MASK]}' tokens to serve as input in our \textit{cloze test} setting. CODE is divided into three splits: the \texttt{train} split has 665,161 bounding boxes in 29,781 images, the \texttt{val} split has 22,061 bounding boxes in 1,000 images, and the \texttt{test} split has 21,641 bounding boxes in 999 images. In total, the CODE dataset has 10,346 unique object names, surpassing the number of object names in any previous detection dataset, such as COCO~\citep{lin2014microsoft} (80 classes) and LVIS~\citep{gupta2019lvis} (1,203 classes).

\noindent \textbf{Data Format.}
Our \benchmarkname benchmark follows the data format of the COCO dataset and includes additional fields to facilitate the evaluation, as shown in Figure~\ref{fig:contextdet_appendix_dataformat}.
The images and annotations used in our new benchmark are based on Flickr 30k~\citep{young2014image} and Flickr30k Entities~\citep{plummer2015flickr30k}.
We tokenize the language caption using the LLM tokenizer and record the related language tokens.
For each object name that appears in the tokens generated by the tokenizer, we track the start and end indices, which will be replaced with the \texttt{[MASK]} token for our contextual cloze test task.

\noindent \textbf{Word Clouds.}
In the contextual cloze test setting, our \benchmarkname dataset consists of 10,346 unique object words that are masked and required to be predicted. Figure~\ref{fig:appendix_word_cloud} presents the word cloud visualizations of object words in our dataset.
We can observe both high-frequency words such as `man' and `woman,' as well as low-frequency words such as `player', `scooty', and `breadstick,' which pose challenges for accurate predictions. Therefore, achieving precise predictions for these object words requires understanding contextual information.

\begin{table*}[t]
\centering
\caption{ \small
Benchmark results of \methodname on our \benchmarkname dataset \texttt{val} set.
We report the classification accuracy (Acc) metric (\%) of \textit{cloze} test and the detection AP metric (\%) of all our three settings for comparisons. The symbol `*' refers to fine-tuning on CODE dataset.
}
\label{tab:code_main_results}
\resizebox{0.8\textwidth}{!}{
\small
\setlength{\tabcolsep}{2pt}
\begin{tabular}{c c|cccc|cc|cc}
   \toprule
   \multirow{2}{*}{\rowNumber{\#}} & \multirow{2}{*}{Method} & \multicolumn{4}{c}{\cellcolor{orange!30}Cloze Test} & \multicolumn{2}{c}{\cellcolor{ForestGreen!30}QA} & \multicolumn{2}{c}{\cellcolor{Cyan!30}Captioning} \\
   ~ & ~ & Acc@1 & Acc@5 & AP@1 & AP@5 & AP@1 & AP@5 & AP@1 & AP@5 \\
    \midrule
    \rowNumber{1} & BLIP-2~\citep{li2023blip2}+GLIP~\citep{li2022grounded} & 40.4 & 69.0 & 9.0 & 19.1 & 5.5 & 13.4 & 5.2 & 7.0 \\
    \rowNumber{2} & LLaVA~\citep{liu2023visual}+GLIP & 41.8 & 70.4 & 8.6 & 18.8 & 5.8 & 13.7 & 4.4 & 7.7 \\
    \rowNumber{3} & LLaVA1.5~\citep{liu2023improved}+GLIP & 42.9 & 71.1 & 9.2 & 19.0 & 6.5 & 14.0 & 5.5 & 7.6 \\
    \rowNumber{4} & LLaVA1.5~\citep{liu2023improved}+GLIP* & 43.2 & 71.5 & 9.1 & 19.4 & 6.3 & 13.9 & 5.8 & 8.0 \\
    \rowNumber{5} & GLIP-2~\citep{zhang2022glipv2} & - & - & - & - & 2.4 & 11.9 & 5.7 & 6.6 \\
    \rowcolor{violet!10}
    \rowNumber{6} & \methodname & \textbf{48.7} & \textbf{73.8} & \textbf{10.2} & \textbf{20.5} & \textbf{8.1} & \textbf{18.2} & \textbf{6.4} & \textbf{9.2} \\
    \bottomrule 
\end{tabular}
}
\end{table*}

\noindent \textbf{Evaluation Details.}
Existing object detection datasets, such as Pascal VOC~\citep{everingham2010pascal}, Microsoft COCO~\citep{lin2014microsoft}, Open Images~\citep{kuznetsova2020open}, LVIS~\citep{gupta2019lvis}, Objects365~\citep{shao2019objects365} and V3Det~\citep{wang2023v3det}, rely on predefined mappings between label IDs and class names for evaluation purposes. For example, the COCO dataset uses a mapping like {(1, person), (2, bicycle), $\dotsc$, (80, toothbrush)} for its 80 classes. As shown in Fig.\ref{fig:contextdet_appendix_evaluation}(a), in order to be classified as true positives, predicted bounding boxes must exhibit both high IoU overlap and identical class IDs to the ground-truth boxes. In certain scenarios, such as zero-shot\citep{bansal2018zero} or open-vocabulary~\citep{zareian2021open} object detection settings, the predefined classes are divided into two separate groups: \emph{base} and \emph{novel}, to evaluate the model's generalization capability. However, these evaluations still rely on the predefined ID-name mappings, while objects with names not included in predefined mappings are impossible.

Human perception does not depend on pre-defined class IDs.
Therefore, for our proposed contextual cloze test task, we have established new evaluation criteria that use object names from human language. In this evaluation, given a masked language expression and the indexes of the masked words, we classify predicted boxes as true positives if they i) exhibit high IoU overlap, ii) share the same meaning, and iii) have an identical masked index as the ground truth boxes. Conversely, predictions are considered false positives. The masked indexes are employed to differentiate cases where multiple objects have the same name but are located at different \texttt{[MASK]} token positions within a sentence. The object names correspond to the most valid English words decoded by the Tokenizer of LLMs.

After defining our name-based criteria as true-positive/false-positive metrics, we could compute the overall Average Precision (AP) metric for evaluation.
We follow the COCO dataset to set the IoU thresholds ranging from 0.5 to 0.95 with a step size of 0.05. The per-name AP is not computed because there are numerous long-tailed infrequent names, of which only a few examples are available for evaluation. 

\noindent \textbf{AP@5 for Top-5 Predictied Names.} 
In some cases, our evaluation metric can be overly stringent, particularly when dealing with numerous synonyms or fine-grained categories that are challenging for annotators to distinguish. Similar challenges have been encountered in previous image classification datasets like ImageNet~\citep{deng2009imagenet}, where the top-5 accuracy metric is used as a supplementary metric to the top-1 accuracy metric. Therefore, we also introduce a supplementary metric called top-5 AP (AP@5), which relaxes the definition of true positives. Under AP@5, if the ground-truth name is among the top-5 predictions, the predictions are considered true positives. In contrast, the AP metric calculated based on the top-1 prediction result is referred to as AP@1 to differentiate it from AP@5.

\noindent \textbf{Implementation Details.}
We modify the famous \textit{pycocotools} package provided in the COCO dataset and create the evaluation script.

\noindent \textbf{Evaluation for Context QA and Context Captioning.}
The contextual understanding performance of the cloze test/captioning/QA settings are highly related. The quantitative evaluation in one setting can readily be converted to another with different text prompts. For example:
\begin{itemize}
\item Cloze Test Prompt: \texttt{A \_\_\_ stands with his bride while holding balloons}. (Expected answer: groom)
\item Captioning Prompt: \texttt{A photo of a \_\_\_ standing with his bride while holding balloons}. (Expected answer: groom)
\item QA Prompt: \texttt{Question: who is standing with the bride while holding balloons in this image? Answer: \_\_\_}. (Expected answer: The groom)
\end{itemize}
All three settings essentially evaluate the contextual understanding of the same scenario but are presented differently. 

\begin{table*}[t]
\centering
\caption{ \small
Benchmark results of \methodname on our \benchmarkname dataset \texttt{val} set with various language models and vision backbones.
We also report the total number of parameters, number of trainable parameters, training time $\rm{T}_{\rm{train}}$ and testing time $\rm{T}_{\rm{test}}$ for efficiency analysis.
}
\label{tab:code_results}
\vspace{-6pt}
\resizebox{0.85\textwidth}{!}{
\small
\setlength{\tabcolsep}{2pt}
\begin{tabular}{c cc|cccc|cc|cc|cc|cc}
   \toprule
   \multirow{2}{*}{\rowNumber{\#}} & Language & Vision & \multicolumn{4}{c}{\cellcolor{orange!30}Cloze Test} & \multicolumn{2}{c}{\cellcolor{ForestGreen!30}QA} & \multicolumn{2}{c}{\cellcolor{Cyan!30}Captioning} & Total & Learnable & $\rm{T}_{\rm{train}}$ & $\rm{T}_{\rm{test}}$ \\
   ~ & Model & Backbone & Acc@1 & Acc@5 & AP@1 & AP@5 & AP@1 & AP@5 & AP@1 & AP@5 & \#Params(M) & \#Params(M) & (s/iter) & (s/iter) \\
    \midrule
    \rowNumber{1} & \multirow{2}{*}{OPT-2.7B} & ResNet50 & 48.7 & 73.8 & 10.2 & 20.5 & 8.1 & 18.2 & 6.4 & 9.2 & 2835 & 183 & 0.437 & 0.224\\
    \rowNumber{2} & ~ & Swin-B & 54.3 & 78.1 & 13.1 & 25.3 & 8.6 & 20.0 & 8.2 & 12.4 & 2893 & 241 & 0.625 & 0.241 \\
    \cmidrule(r){2-15}
    \rowNumber{3} & \multirow{2}{*}{OPT-6.7B} & ResNet50 & 49.2 & 74.5 & 11.1 & 23.4 & 8.0 & 7.5 & 11.6 & 21.3 & 6922 & 263 & 0.448 & 0.248 \\
    \rowNumber{4} & ~ & Swin-B & 54.8 & 78.6 & 13.7 & 26.6 & 9.0 & 23.4 & 9.2 & 14.0 & 6979 & 320  & 0.652 & 0.251 \\
    \bottomrule
\end{tabular}
\vspace{-6pt}
}
\end{table*}

\begin{table*}[t!]
\begin{minipage}{.44\textwidth}
\centering
\small
\caption{ \small
Ablation studies on the impact of using local visual tokens $\vz$.
}
\label{tab:ablation_c}
\vspace{-6pt}
\resizebox{0.7\textwidth}{!}{
\small
\setlength{\tabcolsep}{2pt}
\begin{tabular}{c c | cccc}
\toprule
\rowNumber{\#} & $\vz$ & Acc@1 & Acc@5 & AP@1 & AP@5 \\
\midrule
\rowNumber{1} & \xmark & 30.9 & 57.1 & 4.0 & 13.6 \\
\rowcolor{violet!10}
\rowNumber{2} & \cmark & \textbf{48.7} & \textbf{73.9} & \textbf{10.4} & \textbf{21.6} \\
\bottomrule
\end{tabular}}
\vspace{-6pt}
\end{minipage}
\hspace{+1pt}
\begin{minipage}{.48\textwidth}
\centering
\small
\caption{ \small
Ablation study: varying values of $p$.
}
\label{tab:ablation_p}
\vspace{-6pt}
\resizebox{0.85\textwidth}{!}{
\begin{tabular}{c c | cccc}
\toprule
\rowNumber{\#} & $p$ & Acc@1 & Acc@5 & AP@1 & AP@5 \\
\midrule
\rowNumber{1} & 4 & 48.4 & 73.2 & 10.1 & 20.1 \\
\rowcolor{violet!10}
\rowNumber{2} & 9 & \textbf{48.7} & \textbf{73.9} & \textbf{10.4} & \textbf{21.6} \\
\rowNumber{3} & 16 & 47.5 & 72.9 & 9.9 & 19.4 \\
\bottomrule
\end{tabular}}
\vspace{-6pt}
\end{minipage}
\end{table*}

\section{Experiments}
We present the results of \methodname on different tasks, including (1) our proposed contextual object detection task discussed in Sec.~\ref{sec:results_contextual_cloze_test}, and existing tasks such as (2) open-vocabulary object detection in Sec.~\ref{sec:ov_det}, and (3) referring image segmentation in Sec.~\ref{sec:ref_seg}.

\noindent \textbf{Implementation Details.}
Our proposed method is implemented in PyTorch and all models are trained using a single machine with 4 NVIDIA A100 GPUs. During training, data augmentation techniques are applied including random horizontal flipping with a probability of 0.5 and large-scale jittering~\citep{ghiasi2020simple}. We set the batch size to 8 and train the model for 6 epochs. We use AdamW~\citep{loshchilov2017decoupled} optimizer with a learning rate of $1e^{-4}$ and a weight decay of $0.05$.
For \methodname, we report the results using OPT-2.7B~\citep{zhang2022opt} as the language model and ResNet50~\citep{he2016deep} as the vision backbone.

\subsection{Contextual Object Detection}
\label{sec:results_contextual_cloze_test}

In this section, we report the benchmark results on our proposed CODE Dataset (Sec.~\ref{appendix:dataset_details}).

\noindent \textbf{Evaluation Metrics.}
In our contextual \textit{cloze test} setting, we compute both classification accuracy and detection AP metrics. The accuracy means the percentage of correctly predicted object words. However, evaluating this accuracy poses a challenge due to the presence of numerous synonyms and fine-grained object words in human language, which can be difficult for annotators to distinguish.
This is a problem similar to those faced by previous large vocabulary image-classification datasets, such as ImageNet~\citep{deng2009imagenet}, which use the top-5 accuracy metric as a supplementary metric to the top-1 accuracy.
Consequently, we also adopt both the top-1 accuracy (Acc@1) and the top-5 accuracy (Acc@5) as our evaluation metrics.
For box evaluation, we compute the mean Average Precision (mAP) metric based on the top-1 and top-5 predicted names, which are represented as AP@1 and AP@5.
In evaluation, we compared the object name words rather than pre-defined category IDs, which allows a flexible extension to accommodate a vast human vocabulary.
For our contextual \textit{QA} and \textit{captioning} settings, we obtain quantitative results by modifying the input text prompt formats used in the \textit{cloze test} task.

\noindent \textbf{Baselines.}
Due to previous approaches do not have the generate-then-detect capability, we use a cascade solution to combine existing approaches as our baselines: we first generate captions using BLIP-2~\citep{li2023blip2}, LLaVA~\citep{liu2023visual} or LLaVA 1.5~\citep{liu2023improved}, followed by the representative grounding approach GLIP~\citep{li2022grounded}.
We also select the GLIP-2~\citep{zhang2022glipv2} as our baseline for its compatibility with our contextual \textit{QA} and \textit{captioning} scenarios. However, GLIP-2 is not available for the \textit{cloze test} setting, as complete object names are required for GLIP-2.

\noindent \textbf{Results.}
We provide the benchmark results of \methodname on the CODE dataset in Tab.~\ref{tab:code_main_results}. Our results suggest that our contextual object detection is very challenging: the top-1 AP falls significantly below the performance of previous object detection datasets like COCO, largely attributable to our benchmark encompassing 10,346 unique object names (80 names for COCO). Compared to the cascade solution such as BLIP-2/LLaVA/LLaVA1.5+GLIP, our end-to-end \methodname achieves better performance on all the settings. This is attributed to the hidden embeddings extracted from LLMs with self-attention operators containing the contextual relationship between visual and text tokens, which is crucial for our task that requires contextual understanding ability.

We also compare \methodname with LLaVA 1.5 plus fine-tuning the GLIP detector on the CODE dataset (Row \rowNumber{\#4} in Tab.~\ref{tab:code_main_results}).
However, we found that fine-tuning GLIP does not lead to significant improvements in detection performance. This is because GLIP relies on contrastive learning, which may not effectively adapt to the large number of object categories with subtle differences.
Our observation suggests that the contextual information provided by the LLM tokens is essential for detecting novel objects in a vast vocabulary (\eg, larger than 10k classes in our CODE dataset.)

\begin{table*}[t]
\renewcommand{\arraystretch}{1.1}
\begin{minipage}{.4\textwidth}
\centering
\small
\caption{ \small
Comparison with state-of-the-art open-vocabulary detection methods on OV-COCO.
}
\label{table:results_ovcoco}
\vspace{-6pt}
\resizebox{0.83\textwidth}{!}{%
 \begin{tabular}{l lccc} 
 \toprule
 \rowNumber{\#} & Method & AP$_{50}^{\text{novel}}$ & AP$_{50}^{\text{base}}$ & AP$_{50}$ \\
 \midrule
 \rowNumber{1} & ViLD~\citep{gu2021open} & 27.6 & 59.5 & 51.2 \\
 \rowNumber{2} & OV-DETR~\citep{zang2022open} & 29.4 & 61.0 & 52.7 \\
 \rowNumber{2} & BARON~\citep{wu2023aligning} & 34.0 & 60.4 & 53.5 \\
 \midrule
 \rowcolor{violet!10}
 \rowNumber{4} & \methodname & \textbf{36.8} & \textbf{65.1} & \textbf{57.7}  \\
\bottomrule
\end{tabular}}
\end{minipage}
\hspace{+1pt}
\begin{minipage}{.55\textwidth}
\centering
\small
\caption{ \small
Comparisons with state-of-the-art methods on three referring image segmentation benchmarks in terms of the mIoU metric.
}
\vspace{-6pt}
\resizebox{0.99\textwidth}{!}{%
 \begin{tabular}{ll|ccc|ccc|cc} 
 \toprule
 \multirow{2}{*}{\rowNumber{\#}} & \multirow{2}{*}{Method}  &
 \multicolumn{3}{c|}{\cellcolor{orange!30}RefCOCO} &  
 \multicolumn{3}{c|}{\cellcolor{ForestGreen!30}RefCOCO+} & 
 \multicolumn{2}{c}{\cellcolor{Cyan!30}RefCOCOg}  \\ [0.5ex] 
        ~ & ~ & val & testA & testB & val & testA & testB & val & test  \\
 \midrule
 \rowNumber{1} & RefTR~\citep{muchen2021referring} & 74.34 & 76.77 & 70.87 & 66.75 & 70.58 & 59.40 & 66.63 & 67.39 \\
 \rowNumber{2} & LAVT~\citep{yang2022lavt} & 74.46 & 76.89 & 70.94 & 65.81 & 70.97 & 59.23 & 63.34 & 63.62 \\
 \rowNumber{3} & PolyFormer~\citep{liu2023polyformer} & 75.96 & 77.09 & 73.22 & 70.65 & 74.51 & 64.64 & 69.36 & 69.88 \\
\midrule
\rowcolor{violet!10} \rowNumber{4} & \methodname & \textbf{76.40} & \textbf{77.39} & \textbf{74.16} & \textbf{71.67} & \textbf{75.14} & \textbf{65.52} & \textbf{69.89} & \textbf{70.33}  \\
\bottomrule
\end{tabular}}
\label{table:results_refeseg}
\end{minipage}
\end{table*}

\subsection{Ablation Studies}\label{appendix:ablation_studies}
We investigate the effects of using local visual tokens $\vz$, the associated hyper-parameter $p$ that determines the number of local bins, and the efficiency analysis. 
The experiments are conducted on the CODE \texttt{val} set.

\noindent \textbf{Hyper-Parameter $p$.} As discussed in Section~\ref{sec:visual_encoder}, we have $p$ visual local tokens that serve as prefix inputs for LLM decoding.
In Table~\ref{tab:ablation_p}, we show the effects of using different values for $p$.
We observe that selecting $p=9$ (Row \rowNumber{\#2}) yields the optimal results, making it our default choice.

\noindent \textbf{More Backbones.} We provide the results of \methodname on the CODE dataset in Table~\ref{tab:code_results}.
We first report the results using OPT-2.7B~\cite{zhang2022opt} as the language model and ResNet50~\cite{he2016deep} as the vision backbone (Row \rowNumber{\#1}). Our results suggest that the contextual cloze test task is very challenging: the top-1 AP (AP@1) is just 10.2, which falls significantly below the performance of previous object detection datasets like COCO. Moreover, our study suggests that using more powerful language models and vision backbones can improve performance. When we replace ResNet50 with Swin-B~\cite{liu2021swin} (Row \rowNumber{\#2}), we observe a notable improvement from $10.2$ to $13.1$ in AP@1. In addition, by replacing OPT-2.7B with the larger OPT-6.7B (Row \rowNumber{\#4}), we achieve an even higher AP@1 performance of $13.7$.

\noindent \textbf{LLM without Local Visual Tokens.}
In our contextual cloze test setting, LLM is capable of making predictions even without the presence of the local visual token input $\vz$. However, upon analyzing the results presented in Table~\ref{tab:ablation_c}, we observe a significant performance drop. For example, the top-1 accuracy drops around 20 percent from 48.7 to 30.9 (\%). This observation emphasizes the crucial role of adding visual local tokens in our method for contextual understanding. We also observe that the value of language modeling loss $\mathcal{L}_{\text{lm}}$ barely decreases in the absence of $\vz$. This observation is because computing the language modeling loss $\mathcal{L}_{\text{lm}}$ is closely related to generating the next text tokens, which relies heavily on the visual tokens $z$. Without the visual tokens $z$, the model is unable to effectively generate text tokens that accurately describe the visual content, resulting in a stagnant language modeling loss.

\noindent \textbf{Efficiency Analysis.}
The majority of parameters in our model, including the LLM component, are frozen, resulting in a small percentage of learnable parameters. As shown in Table~\ref{tab:code_main_results} Row \rowNumber{\#1}, when employing OPT-2.7B and the ResNet50 backbone, only 6.4\% (183 out of 2,835) of parameters are trainable. Our design does not impose a significant computational burden and can be easily reproduced.

\begin{figure*}[t]
    \centering
    \small
    \includegraphics[width=.86\textwidth]{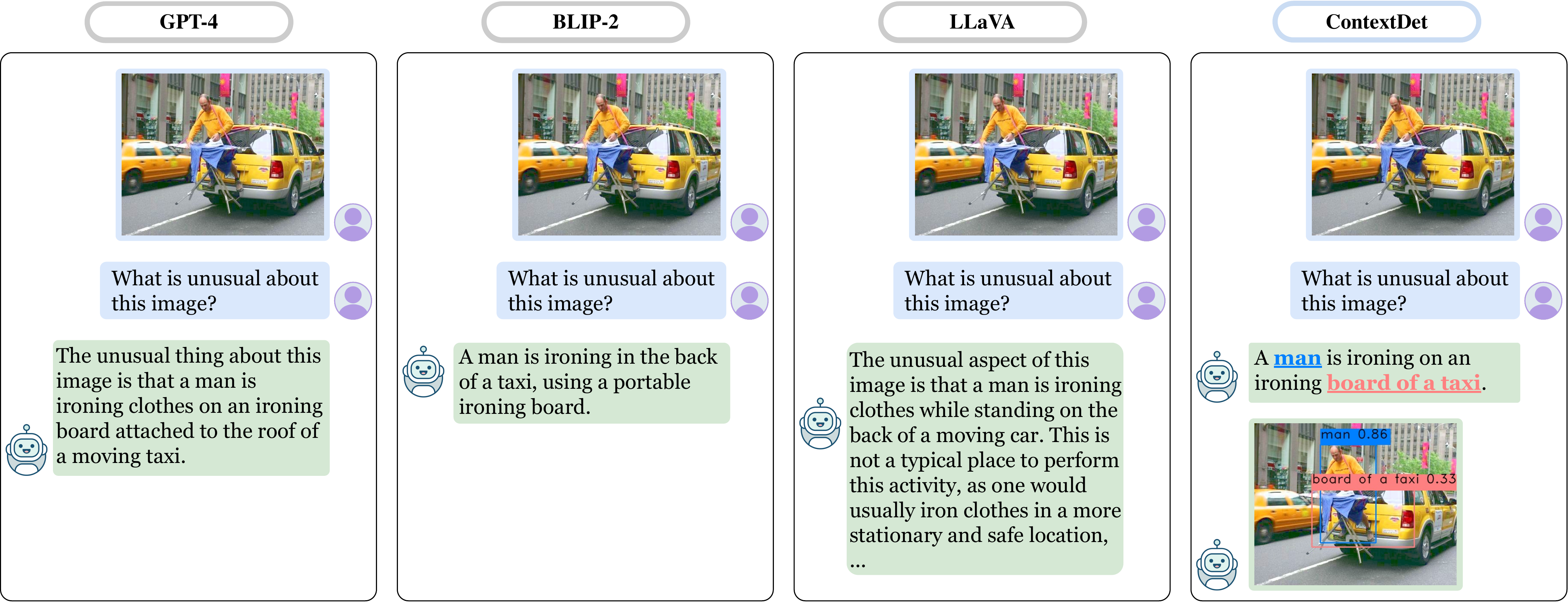}
    \caption{ \small
     Qualitative examples comparing \methodname with existing Multimodal Language Models (MLLMs), including GPT-4~\citep{openai2023gpt4}, BLIP-2~\citep{li2023blip2}, and LLaVA~\citep{liu2023visual}.
     Our method predicts related bounding boxes for the object names mentioned in the text outputs,
     (\eg, `\textcolor{ForestGreen!90}{man}', `\textcolor{orange!90}{board of a taxi}'),
     enabling a more comprehensive interpretation for visual-language tasks and paving the way for broader application areas.
    }
    \label{fig:contextdet_vs_mllm}
    \vspace{-6pt}
\end{figure*}

\subsection{Open-Vocabulary Object Detection}
\label{sec:ov_det}
We demonstrate that our proposed \methodname can also be applied to the open-vocabulary object detection task, aiming to evaluate the generalization ability. Following previous works~\citep{bansal2018zero,zareian2021open}, we use the OV-COCO benchmark and divide 65 categories as the split of base/novel (48/17) classes. The model is trained on the base classes only but evaluated on the novel classes (unavailable during model training). We measure the performance with the Average Precision (AP) metric on the base, novel, and all classes.

To adapt \methodname into the open-vocabulary setting, we ask questions like `\texttt{Does the [CLASS] appear in this picture?}' for every class including base and novel classes. If MLLM responds with a positive answer `\texttt{Yes}', we take the latent embedding $\ve$ of the corresponding class name as a conditional input for our visual decoder (Sec.~\ref{sec:visual_decoder}).
We compare \methodname with selected baseline methods including the state-of-the-art method BARON~\citep{wu2023aligning} in Table.~\ref{table:results_ovcoco}.
We observe that \methodname significantly outperforms BARON by large margins of 2.8\%, 4.7\%, and 4.2\% on the novel, base, and all sets, respectively.
All the baseline methods rely on prior knowledge from the vision-language model CLIP.
In contrast, our \methodname uses MLLM to detect novel objects.
The results show that MLLM trained on web-scale datasets has strong generalizability that could benefit the open-vocabulary task.

\textbf{Efficiency Analysis.} The training and test time (sec/iter) performances of OV-DETR are 0.47/0.63 and 0.54/0.25 for our approach. While our approach increases the training time due to a higher number of parameters of LLMs, it proves to be more efficient during testing. This efficiency is attributed to the fact that while OV-DETR conditions on all classes, ContextDet specifically conditions classes present in the image by asking the questions.

\subsection{Referring Image Segmentation}\label{sec:ref_seg}
Our \methodname is not limited to object detection and can be extended to the image segmentation task, in which the goal is to assign a pixel-level label to each pixel in the input image.
To adapt our \methodname framework for segmentation, we introduce an extra pixel-level segmentation head that takes the full visual tokens $\vc$ as inputs.
To train the segmentation model, we use a pixel-wise cross-entropy loss $\mathcal{L}_{\text{mask}}$ and Dice loss $\mathcal{L}_{\text{dice}}$, where ground-truth labels are pixel-level masks for matched objects in an image.

We choose the referring image segmentation task as a representative benchmark to evaluate the segmentation performance of \methodname.
The referring image segmentation task aims to segment regions described by fine-grained input language query. Language queries will act as conditional inputs for the visual decoder in \methodname.
We use three commonly-used datasets: RefCOCO~\citep{yu2016modeling}, RefCOCO+~\citep{yu2016modeling} and RefCOCOg~\citep{nagaraja2016modeling}.
On RefCOCO and RefCOCO+, we follow the default training/validation/testA/testB data split in Yu \textit{et al.} \citep{yu2016modeling}.
For RefCOCOg, we use the RefCOCO-umd splits~\citep{nagaraja2016modeling}.
We report the mean Intersection over Union (mIoU), which is calculated by averaging the IoU scores across all test samples.
We compare \methodname with some state-of-the-art methods in Table.~\ref{table:results_refeseg}. \methodname achieves better results with mIoU gains of 0.63\% and 0.45\% on the validation/test splits over PolyFormer~\citep{liu2023polyformer}.

\subsection{Standard Object Detection}
\begin{table}[t!]
\begin{minipage}{.44\textwidth}
\centering
\caption{\small
Comparison the standard object detection results on the COCO validation dataset.
}
\label{tab:coco}
\vspace{-6pt}
\resizebox{0.75\textwidth}{!}{
\small
\setlength{\tabcolsep}{2pt}
\begin{tabular}{c c | ccc ccc}
\toprule
\rowNumber{\#} & Model & AP & $AP_{50}$ & $AP_{75}$ & $AP_{s}$ & $AP_{m}$ & $AP_{l}$ \\
\midrule
\rowNumber{1} & DETR & 42.0 & 62.4 & 44.2 & 20.5 & 45.8 & 61.1 \\
\rowNumber{2} & Deformable DETR & 46.2 & 65.2 & 50.0 & 28.8 & 49.2 & 61.7 \\
\rowcolor{violet!10}
\rowNumber{3} & ContextDET & 43.4 & 62.9 & 47.4 & 26.0 & 46.9 & 56.8 \\
\bottomrule
\end{tabular}}
\vspace{-6pt}
\end{minipage}
\end{table}
We further evaluate the effectiveness of ContextDET on a standard object detection task using the COCO benchmark. By applying the `captioning' setting of ContextDET and implementing post-processing adjustments such as filtering out irrelevant object categories and confident thresholding, the results are presented in Tab.~\ref{tab:coco}. While our performance does not match specialized object detection methods like Deformable DETR, ContextDET demonstrates strengths in detecting small objects, which we attribute to the benefits of contextual information in disambiguating regions containing small objects.

\subsection{Qualitative Results}
\label{sec:qualitative_results}

\begin{figure*}[t]
    \centering
    \small
    \includegraphics[width=.9\textwidth]{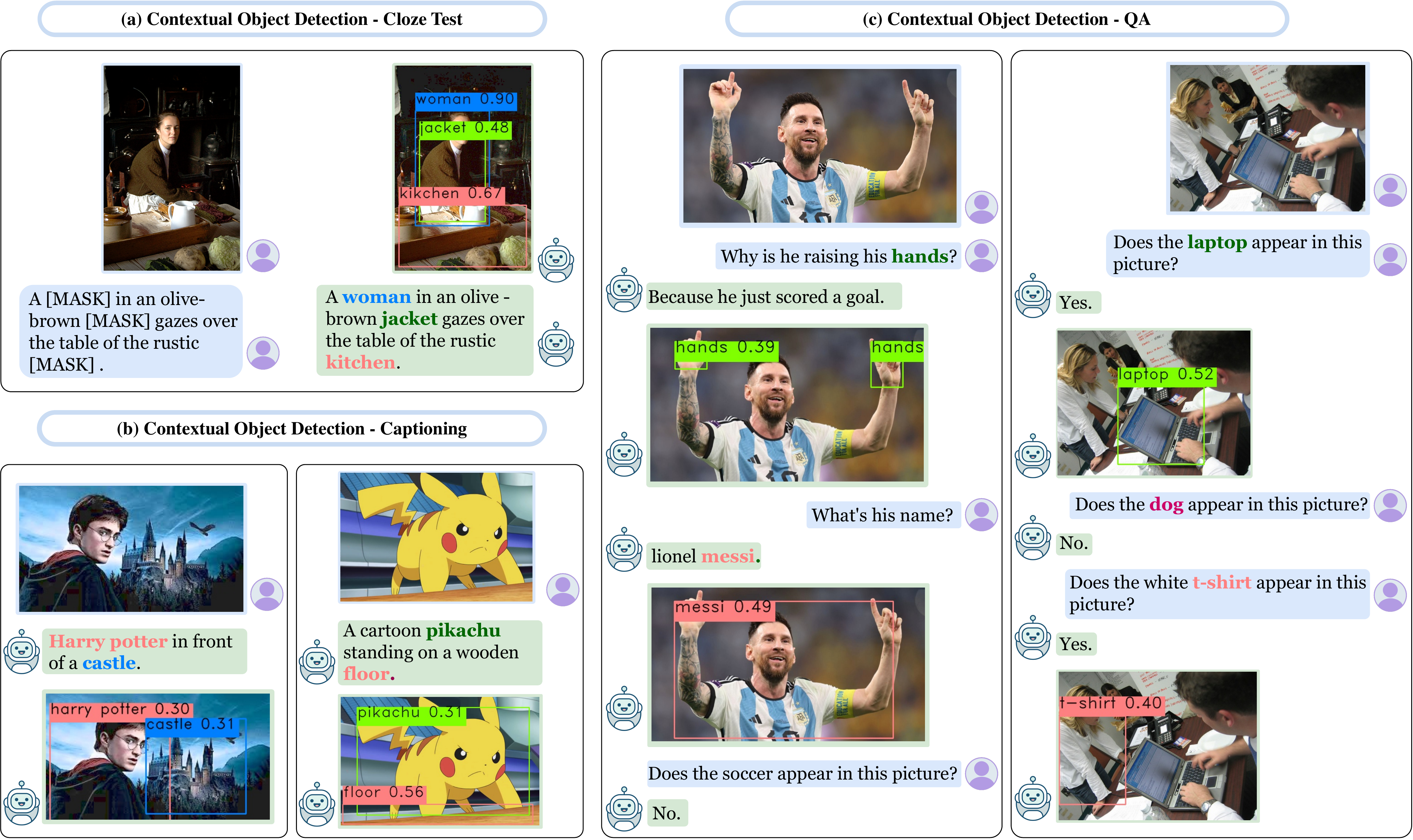}
    \caption{ \small
     Qualitative examples predicted by \methodname in our three contextual object detection settings include (a) cloze test, (b) captioning, and (c) question answering.
    The `harry potter', `pikachu', and `messi' are novel names that are not annotated in the CODE training set.
    \methodname shows plausible contextual understanding and generalization abilities.
    }
    \label{fig:contextdet_vis}
\end{figure*}

\begin{figure*}
    \centering
    \includegraphics[width=0.85\textwidth]{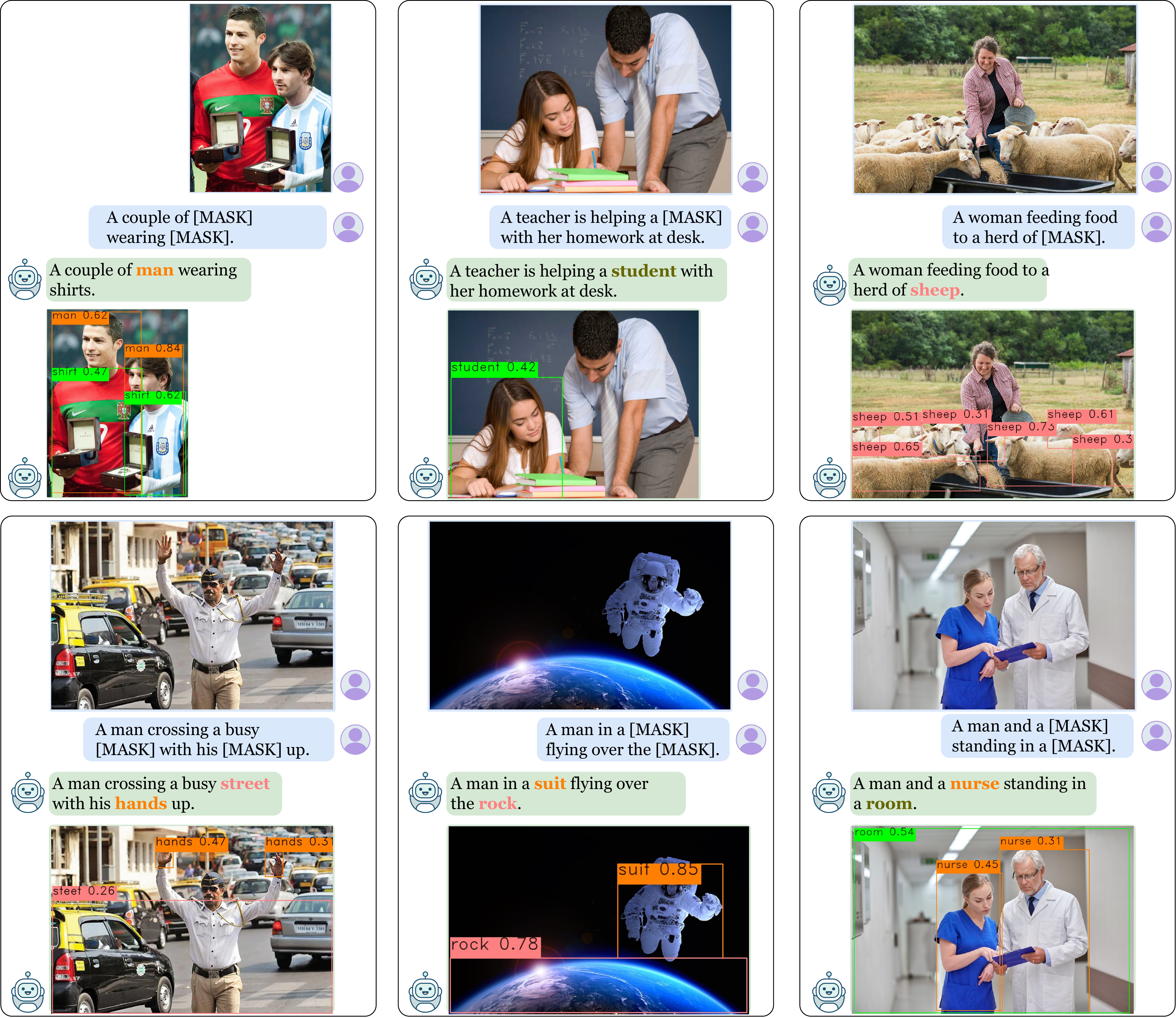}
    \caption{
    \small
    Qualitative examples of the \textbf{contextual cloze test}.
    }
    \label{fig:contextdet_appendix_vis_cloze}
    \vspace{+2pt}
    \centering
    \includegraphics[width=0.85\textwidth]{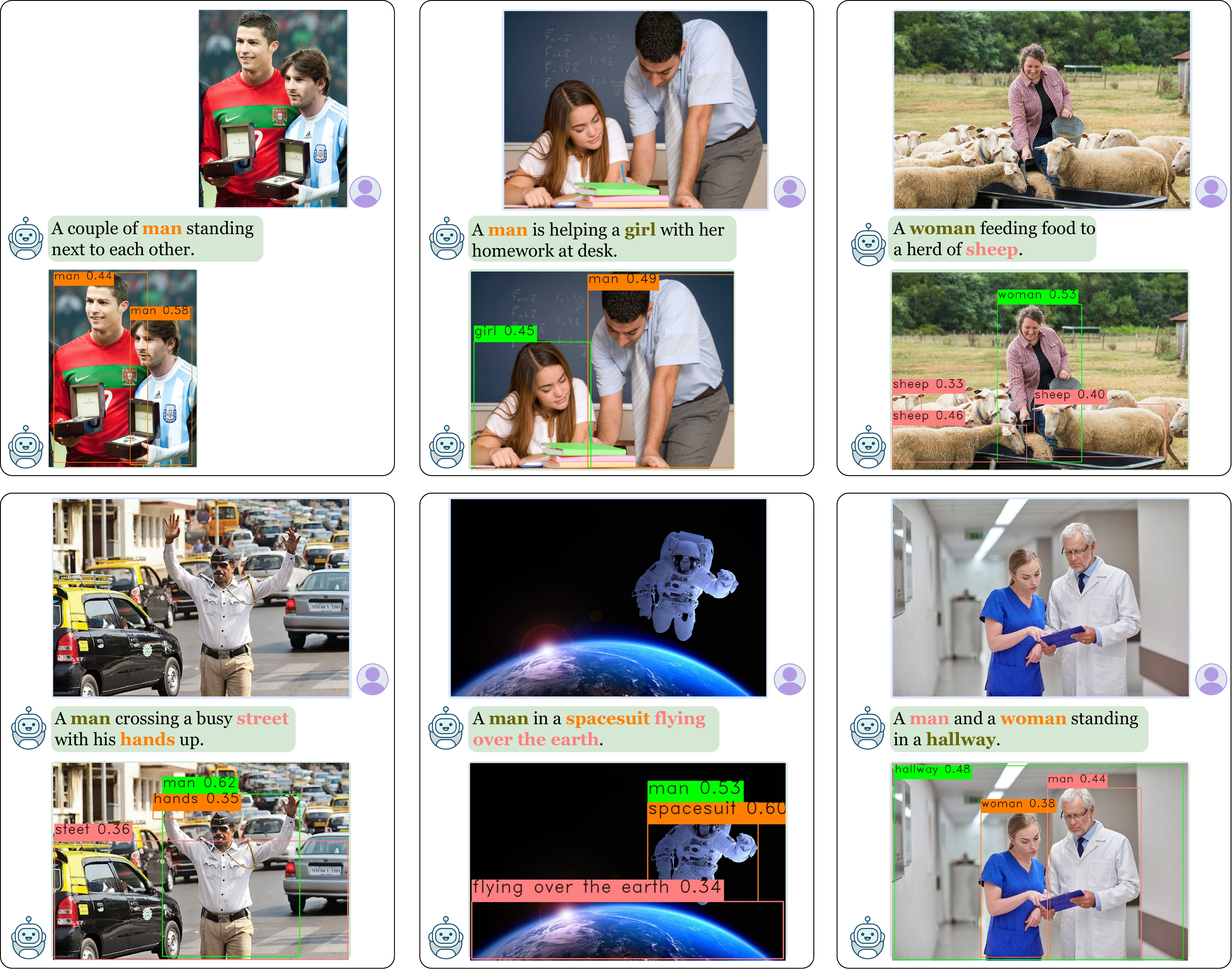}
    \caption{
    \small
    Qualitative examples of the \textbf{contextual captioning}.
    }
    \label{fig:contextdet_appendix_vis_captioning}
\end{figure*}

\begin{figure*}
    \centering
    \includegraphics[width=0.85\textwidth]{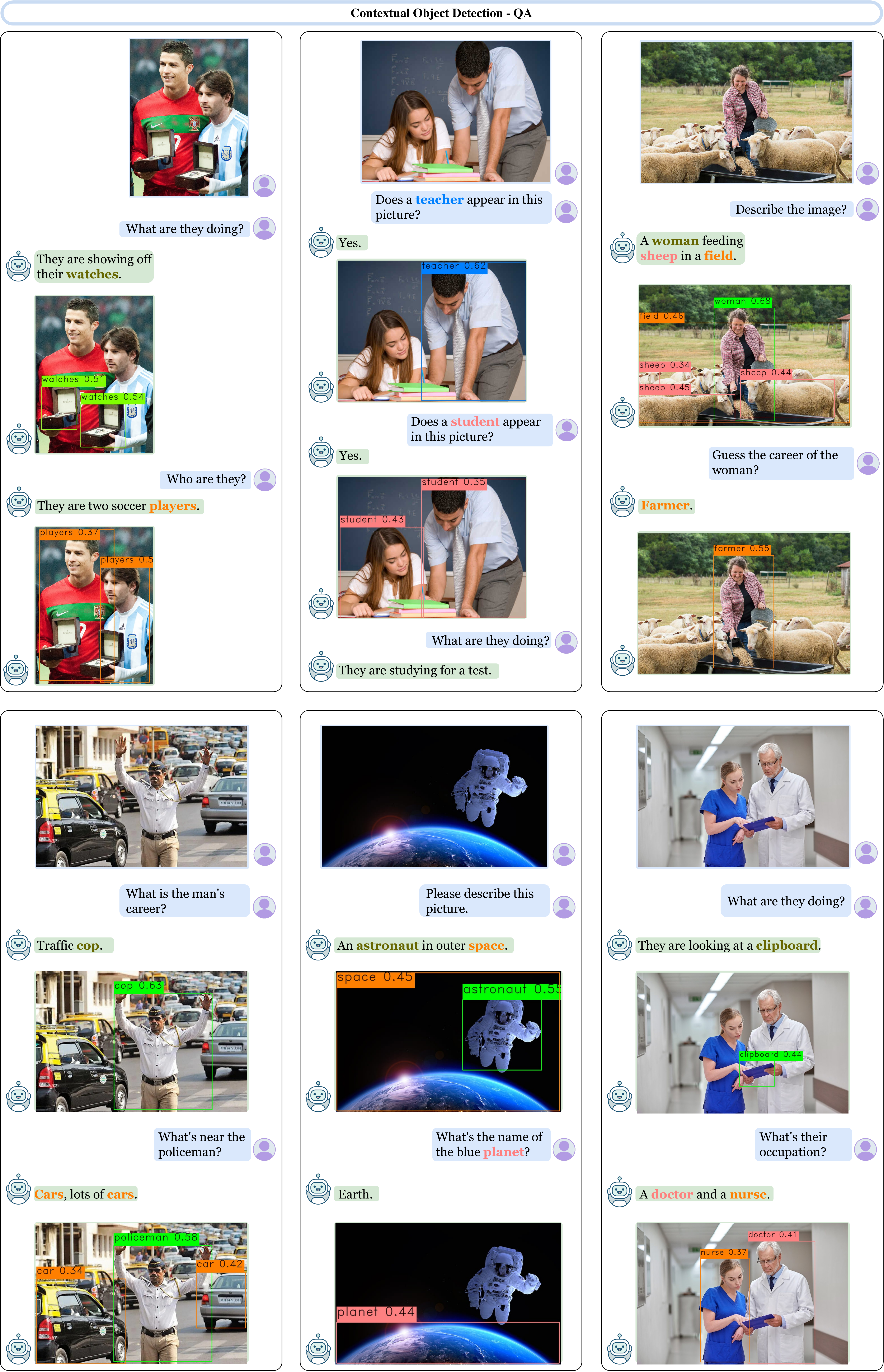}
    \caption{
    \small
    Qualitative examples of the \textbf{contextual QA}.
    }
    \label{fig:contextdet_appendix_vis_qa}
\end{figure*}

Besides the quantitative evaluation on the CODE benchmark, we further qualitatively evaluate \methodname using more diverse images and objects, as shown in Fig.~\ref{fig:contextdet_vis}.
We observe the capacity of \methodname for complex contextual understanding and generalization to open-world names. For example, as illustrated in Fig.~\ref{fig:contextdet_vis} (a), \methodname can reasonably infer the object names to fill the masked tokens, and accurately connect the object names with bounding boxes.  Moreover, \methodname is capable of predicting the names and locations of open-world concepts (\eg, `Harry Potter', `Pikachu', `Messi'), which are difficult to detect using previous close-set object detectors. Finally, in Fig.~\ref{fig:contextdet_vis} (c), we show that \methodname can engage in multi-round question-answering conversations, and predict the bounding boxes of objects mentioned in the dialog history.

We further provide more qualitative results predicted by \methodname in the contextual cloze test (Figure~\ref{fig:contextdet_appendix_vis_cloze}), contextual captioning (Figure~\ref{fig:contextdet_appendix_vis_captioning}), and contextual QA settings~(Figure~\ref{fig:contextdet_appendix_vis_qa}). The selected images are sourced randomly from the web and are not included in the training data. We observe that \methodname effectively predicts contextual object words, including terms like `teacher', `student', `doctor', and `nurse', along with their corresponding bounding boxes. In addition, we find some failure cases. For instance, the predicted object words may be incorrect, particularly for less common terms like `earth'.
Our \methodname is less robust when it comes to occluded objects, such as `sheep'. We aim to address these limitations in future research.

\noindent \textbf{Comparison with MLLMs.}
We present some visual examples in Fig.~\ref{fig:contextdet_vs_mllm} and compare our \methodname with some popular MLLMs like GPT-4~\citep{openai2023gpt4}.
Existing MLLMs can only generate textual outputs while our \methodname pushes the boundaries further by providing bounding boxes of objects of interest. In particular, our method allows fine-grained localization of objects of interest specified in the text input, which offers a higher degree of interpretability for vision-language models. Broadly speaking, our method offers new possibilities for various applications requiring both object localization and conversational interaction, \eg, AR/VR systems and robotics.

\section{Conclusion}

Although recent MLLMs have demonstrated remarkable abilities in vision-language tasks such as question-answering, their potential in other perception tasks remains largely unexplored. Our \methodname highlights the significant potential of MLLMs in diverse perception tasks, such as the proposed contextual object detection task, which predicts precise object names and their locations in images for human-AI interaction. To train our model, we needed to associate object words of bounding boxes with language captions, incurring a high annotation cost. Consequently, we used less training data compared to previous MLLM papers, which may limit our final performance. In future work, we plan to explore the use of semi-supervised or weakly-supervised learning techniques to reduce annotation costs. Additionally, apart from their contextual understanding ability, we believe that other abilities of MLLMs remain underexplored for downstream tasks, such as their interactive ability for instruction tuning. For instance, can MLLMs be utilized to post-process detection outputs based on human language instructions? By providing instructions such as ``shift the predicted box slightly to the left,'' ``remove the redundant overlapped box,'' or ``correct the predicted class from eagle to falcon,'' can MLLMs adjust the predictions accordingly to meet our expectations? We hope the insights presented in this paper could inspire further research into adapting MLLMs to revolutionize more computer vision tasks.

\noindent \textbf{Data Availability Statements} The datasets analyzed during this study are all publicly available for research purposes.

\noindent \textbf{Acknowledgement} This study is supported under the RIE2020 Industry Alignment Fund Industry Collaboration Projects (IAF-ICP) Funding Initiative, as well as cash and in-kind contribution from the industry partner(s). It is also partly supported by the NTU NAP grant and Singapore MOE AcRF Tier 2 (MOE-T2EP20120-0001).


%
%

\bibliographystyle{spbasic}      
\bibliography{ref}   

\begin{thebibliography}{87}
\providecommand{\natexlab}[1]{#1}
\providecommand{\url}[1]{{#1}}
\providecommand{\urlprefix}{URL }
\expandafter\ifx\csname urlstyle\endcsname\relax
  \providecommand{\doi}[1]{DOI~\discretionary{}{}{}#1}\else
  \providecommand{\doi}{DOI~\discretionary{}{}{}\begingroup \urlstyle{rm}\Url}\fi
\providecommand{\eprint}[2][]{\url{#2}}

\bibitem[{Alayrac et~al.(2022)Alayrac, Donahue, Luc, Miech, Barr, Hasson, Lenc, Mensch, Millican, Reynolds, Ring, Rutherford, Cabi, Han, Gong, Samangooei, Monteiro, Menick, Borgeaud, Brock, Nematzadeh, Sharifzadeh, Binkowski, Barreira, Vinyals, Zisserman, and Simonyan}]{Alayrac2022FlamingoAV}
Alayrac JB, Donahue J, Luc P, Miech A, Barr I, Hasson Y, Lenc K, Mensch A, Millican K, Reynolds M, Ring R, Rutherford E, Cabi S, Han T, Gong Z, Samangooei S, Monteiro M, Menick J, Borgeaud S, Brock A, Nematzadeh A, Sharifzadeh S, Binkowski M, Barreira R, Vinyals O, Zisserman A, Simonyan K (2022) Flamingo: a visual language model for few-shot learning. In: NeurIPS

\bibitem[{Anderson et~al.(2018)Anderson, He, Buehler, Teney, Johnson, Gould, and Zhang}]{anderson2018bottom}
Anderson P, He X, Buehler C, Teney D, Johnson M, Gould S, Zhang L (2018) Bottom-up and top-down attention for image captioning and visual question answering. In: CVPR

\bibitem[{Antol et~al.(2015)Antol, Agrawal, Lu, Mitchell, Batra, Zitnick, and Parikh}]{antol2015vqa}
Antol S, Agrawal A, Lu J, Mitchell M, Batra D, Zitnick CL, Parikh D (2015) {VQA}: Visual question answering. In: ICCV

\bibitem[{Bansal et~al.(2018)Bansal, Sikka, Sharma, Chellappa, and Divakaran}]{bansal2018zero}
Bansal A, Sikka K, Sharma G, Chellappa R, Divakaran A (2018) Zero-shot object detection. In: ECCV

\bibitem[{Brown et~al.(2020)Brown, Mann, Ryder, Subbiah, Kaplan, Dhariwal, Neelakantan, Shyam, Sastry, Askell et~al.}]{brown2020language}
Brown T, Mann B, Ryder N, Subbiah M, Kaplan JD, Dhariwal P, Neelakantan A, Shyam P, Sastry G, Askell A, et~al. (2020) Language models are few-shot learners. In: NeurIPS

\bibitem[{Carion et~al.(2020)Carion, Massa, Synnaeve, Usunier, Kirillov, and Zagoruyko}]{carion2020end}
Carion N, Massa F, Synnaeve G, Usunier N, Kirillov A, Zagoruyko S (2020) End-to-end object detection with transformers. In: ECCV

\bibitem[{Chen et~al.(2022{\natexlab{a}})Chen, Guo, Yi, Li, and Elhoseiny}]{chen2022visualgpt}
Chen J, Guo H, Yi K, Li B, Elhoseiny M (2022{\natexlab{a}}) {VisualGPT}: Data-efficient adaptation of pretrained language models for image captioning. In: CVPR

\bibitem[{Chen et~al.(2017)Chen, Zhang, Xiao, Nie, Shao, Liu, and Chua}]{chen2017sca}
Chen L, Zhang H, Xiao J, Nie L, Shao J, Liu W, Chua TS (2017) {SCA-CNN}: Spatial and channel-wise attention in convolutional networks for image captioning. In: CVPR

\bibitem[{Chen et~al.(2022{\natexlab{b}})Chen, Saxena, Li, Fleet, and Hinton}]{chen2021pix2seq}
Chen T, Saxena S, Li L, Fleet DJ, Hinton G (2022{\natexlab{b}}) {Pix2Seq}: A language modeling framework for object detection. In: ICLR

\bibitem[{Chen et~al.(2018)Chen, Huang, and Tao}]{chen2018context}
Chen Z, Huang S, Tao D (2018) Context refinement for object detection. In: ECCV

\bibitem[{Chowdhery et~al.(2022)Chowdhery, Narang, Devlin, Bosma, Mishra, Roberts, Barham, Chung, Sutton, Gehrmann et~al.}]{chowdhery2022palm}
Chowdhery A, Narang S, Devlin J, Bosma M, Mishra G, Roberts A, Barham P, Chung HW, Sutton C, Gehrmann S, et~al. (2022) {PaLM}: Scaling language modeling with pathways. arXiv preprint arXiv:220402311

\bibitem[{Dai et~al.(2023)Dai, Lang, Zeng, Huang, and Li}]{dai2023exploring}
Dai Y, Lang H, Zeng K, Huang F, Li Y (2023) Exploring large language models for multi-modal out-of-distribution detection. arXiv preprint arXiv:231008027

\bibitem[{Deng et~al.(2009)Deng, Dong, Socher, Li, Li, and Fei-Fei}]{deng2009imagenet}
Deng J, Dong W, Socher R, Li LJ, Li K, Fei-Fei L (2009) {ImageNet}: A large-scale hierarchical image database. In: CVPR

\bibitem[{Divvala et~al.(2009)Divvala, Hoiem, Hays, Efros, and Hebert}]{divvala2009empirical}
Divvala SK, Hoiem D, Hays JH, Efros AA, Hebert M (2009) An empirical study of context in object detection. In: CVPR

\bibitem[{Dosovitskiy et~al.(2021)Dosovitskiy, Beyer, Kolesnikov, Weissenborn, Zhai, Unterthiner, Dehghani, Minderer, Heigold, Gelly et~al.}]{dosovitskiy2021image}
Dosovitskiy A, Beyer L, Kolesnikov A, Weissenborn D, Zhai X, Unterthiner T, Dehghani M, Minderer M, Heigold G, Gelly S, et~al. (2021) An image is worth 16x16 words: Transformers for image recognition at scale. In: ICLR

\bibitem[{Driess et~al.(2023)Driess, Xia, Sajjadi, Lynch, Chowdhery, Ichter, Wahid, Tompson, Vuong, Yu, Huang, Chebotar, Sermanet, Duckworth, Levine, Vanhoucke, Hausman, Toussaint, Greff, Zeng, Mordatch, and Florence}]{driess2023palme}
Driess D, Xia F, Sajjadi MSM, Lynch C, Chowdhery A, Ichter B, Wahid A, Tompson J, Vuong Q, Yu T, Huang W, Chebotar Y, Sermanet P, Duckworth D, Levine S, Vanhoucke V, Hausman K, Toussaint M, Greff K, Zeng A, Mordatch I, Florence P (2023) {PaLM-E}: An embodied multimodal language model. arXiv preprint arXiv:230303378

\bibitem[{Du et~al.(2022)Du, Wei, Zhang, Shi, Gao, and Li}]{du2022learning}
Du Y, Wei F, Zhang Z, Shi M, Gao Y, Li G (2022) Learning to prompt for open-vocabulary object detection with vision-language model. In: CVPR

\bibitem[{Everingham et~al.(2010)Everingham, Van~Gool, Williams, Winn, and Zisserman}]{everingham2010pascal}
Everingham M, Van~Gool L, Williams CK, Winn J, Zisserman A (2010) {The PASCAL Visual Object Classes (VOC) Challenge}. IJCV

\bibitem[{Ghiasi et~al.(2021)Ghiasi, Cui, Srinivas, Qian, Lin, Cubuk, Le, and Zoph}]{ghiasi2020simple}
Ghiasi G, Cui Y, Srinivas A, Qian R, Lin TY, Cubuk ED, Le QV, Zoph B (2021) Simple copy-paste is a strong data augmentation method for instance segmentation. In: CVPR

\bibitem[{Goyal et~al.(2017)Goyal, Khot, Summers-Stay, Batra, and Parikh}]{goyal2017making}
Goyal Y, Khot T, Summers-Stay D, Batra D, Parikh D (2017) Making the v in vqa matter: Elevating the role of image understanding in visual question answering. In: CVPR

\bibitem[{Gu et~al.(2022)Gu, Lin, Kuo, and Cui}]{gu2021open}
Gu X, Lin TY, Kuo W, Cui Y (2022) Open-vocabulary object detection via vision and language knowledge distillation. In: ICLR

\bibitem[{Gupta et~al.(2019)Gupta, Dollar, and Girshick}]{gupta2019lvis}
Gupta A, Dollar P, Girshick R (2019) {LVIS}: A dataset for large vocabulary instance segmentation. In: CVPR

\bibitem[{He et~al.(2016)He, Zhang, Ren, and Sun}]{he2016deep}
He K, Zhang X, Ren S, Sun J (2016) Deep residual learning for image recognition. In: CVPR

\bibitem[{He et~al.(2017)He, Gkioxari, Doll{\'a}r, and Girshick}]{he2017mask}
He K, Gkioxari G, Doll{\'a}r P, Girshick R (2017) {Mask R-CNN}. In: CVPR

\bibitem[{Hochreiter and Schmidhuber(1997)}]{hochreiter1997long}
Hochreiter S, Schmidhuber J (1997) Long short-term memory. Neural computation

\bibitem[{Huang et~al.(2023)Huang, Dong, Wang, Hao, Singhal, Ma, Lv, Cui, Mohammed, Liu et~al.}]{huang2023language}
Huang S, Dong L, Wang W, Hao Y, Singhal S, Ma S, Lv T, Cui L, Mohammed OK, Liu Q, et~al. (2023) Language is not all you need: Aligning perception with language models. arXiv preprint arXiv:230214045

\bibitem[{HuggingFace(2024)}]{huggingface}
HuggingFace (2024) Huggingface. \urlprefix\url{https://huggingface.co/}, \url{https://huggingface.co/}

\bibitem[{Jia et~al.(2021)Jia, Yang, Xia, Chen, Parekh, Pham, Le, Sung, Li, and Duerig}]{jia2021scaling}
Jia C, Yang Y, Xia Y, Chen YT, Parekh Z, Pham H, Le Q, Sung YH, Li Z, Duerig T (2021) Scaling up visual and vision-language representation learning with noisy text supervision. In: ICML

\bibitem[{Kamath et~al.(2021)Kamath, Singh, LeCun, Synnaeve, Misra, and Carion}]{kamath2021mdetr}
Kamath A, Singh M, LeCun Y, Synnaeve G, Misra I, Carion N (2021) Mdetr-modulated detection for end-to-end multi-modal understanding. In: CVPR

\bibitem[{Karpathy and Fei-Fei(2015)}]{karpathy2015deep}
Karpathy A, Fei-Fei L (2015) Deep visual-semantic alignments for generating image descriptions. In: CVPR

\bibitem[{Kirillov et~al.(2023)Kirillov, Mintun, Ravi, Mao, Rolland, Gustafson, Xiao, Whitehead, Berg, Lo et~al.}]{kirillov2023segment}
Kirillov A, Mintun E, Ravi N, Mao H, Rolland C, Gustafson L, Xiao T, Whitehead S, Berg AC, Lo WY, et~al. (2023) Segment anything. arXiv preprint arXiv:230402643

\bibitem[{Koh et~al.(2023)Koh, Salakhutdinov, and Fried}]{koh2023grounding}
Koh JY, Salakhutdinov R, Fried D (2023) Grounding language models to images for multimodal generation. arXiv preprint arXiv:230113823

\bibitem[{Kuo et~al.(2022)Kuo, Cui, Gu, Piergiovanni, and Angelova}]{kuo2022f}
Kuo W, Cui Y, Gu X, Piergiovanni A, Angelova A (2022) {F-VLM}: Open-vocabulary object detection upon frozen vision and language models. arXiv preprint arXiv:220915639

\bibitem[{Kuznetsova et~al.(2020)Kuznetsova, Rom, Alldrin, Uijlings, Krasin, Pont-Tuset, Kamali, Popov, Malloci, Kolesnikov et~al.}]{kuznetsova2020open}
Kuznetsova A, Rom H, Alldrin N, Uijlings J, Krasin I, Pont-Tuset J, Kamali S, Popov S, Malloci M, Kolesnikov A, et~al. (2020) The open images dataset v4. IJCV

\bibitem[{Law and Deng(2018)}]{law2018cornernet}
Law H, Deng J (2018) {CornerNet}: Detecting objects as paired keypoints. In: ECCV

\bibitem[{Li et~al.(2023)Li, Li, Savarese, and Hoi}]{li2023blip2}
Li J, Li D, Savarese S, Hoi S (2023) {BLIP-2}: Bootstrapping language-image pre-training with frozen image encoders and large language models. arXiv preprint arXiv:230112597

\bibitem[{Li et~al.(2022)Li, Zhang, Zhang, Yang, Li, Zhong, Wang, Yuan, Zhang, Hwang et~al.}]{li2022grounded}
Li LH, Zhang P, Zhang H, Yang J, Li C, Zhong Y, Wang L, Yuan L, Zhang L, Hwang JN, et~al. (2022) Grounded language-image pre-training. In: CVPR

\bibitem[{Lin et~al.(2014)Lin, Maire, Belongie, Hays, Perona, Ramanan, Doll{\'a}r, and Zitnick}]{lin2014microsoft}
Lin TY, Maire M, Belongie S, Hays J, Perona P, Ramanan D, Doll{\'a}r P, Zitnick CL (2014) {Microsoft COCO}: Common objects in context. In: ECCV

\bibitem[{Liu et~al.(2023{\natexlab{a}})Liu, Li, Li, and Lee}]{liu2023improved}
Liu H, Li C, Li Y, Lee YJ (2023{\natexlab{a}}) Improved baselines with visual instruction tuning. arXiv preprint arXiv:231003744

\bibitem[{Liu et~al.(2023{\natexlab{b}})Liu, Li, Wu, and Lee}]{liu2023visual}
Liu H, Li C, Wu Q, Lee YJ (2023{\natexlab{b}}) Visual instruction tuning. arXiv preprint arXiv:230408485

\bibitem[{Liu et~al.(2023{\natexlab{c}})Liu, Ding, Cai, Zhang, Satzoda, Mahadevan, and Manmatha}]{liu2023polyformer}
Liu J, Ding H, Cai Z, Zhang Y, Satzoda RK, Mahadevan V, Manmatha R (2023{\natexlab{c}}) {PolyFormer}: Referring image segmentation as sequential polygon generation. In: CVPR

\bibitem[{Liu et~al.(2023{\natexlab{d}})Liu, Zeng, Ren, Li, Zhang, Yang, Li, Yang, Su, Zhu et~al.}]{liu2023grounding}
Liu S, Zeng Z, Ren T, Li F, Zhang H, Yang J, Li C, Yang J, Su H, Zhu J, et~al. (2023{\natexlab{d}}) {Grounding DINO}: Marrying dino with grounded pre-training for open-set object detection. arXiv preprint arXiv:230305499

\bibitem[{Liu et~al.(2016)Liu, Anguelov, Erhan, Szegedy, Reed, Fu, and Berg}]{liu2016ssd}
Liu W, Anguelov D, Erhan D, Szegedy C, Reed S, Fu CY, Berg AC (2016) {SSD}: Single shot multibox detector. In: ECCV

\bibitem[{Liu et~al.(2021)Liu, Lin, Cao, Hu, Wei, Zhang, Lin, and Guo}]{liu2021swin}
Liu Z, Lin Y, Cao Y, Hu H, Wei Y, Zhang Z, Lin S, Guo B (2021) {Swin Transformer}: Hierarchical vision transformer using shifted windows. In: CVPR

\bibitem[{Loshchilov and Hutter(2019)}]{loshchilov2017decoupled}
Loshchilov I, Hutter F (2019) Decoupled weight decay regularization. In: ICLR

\bibitem[{Mokady et~al.(2021)Mokady, Hertz, and Bermano}]{mokady2021clipcap}
Mokady R, Hertz A, Bermano AH (2021) {ClipCap}: Clip prefix for image captioning. arXiv preprint arXiv:211109734

\bibitem[{Mottaghi et~al.(2014)Mottaghi, Chen, Liu, Cho, Lee, Fidler, Urtasun, and Yuille}]{mottaghi2014role}
Mottaghi R, Chen X, Liu X, Cho NG, Lee SW, Fidler S, Urtasun R, Yuille A (2014) The role of context for object detection and semantic segmentation in the wild. In: CVPR

\bibitem[{Muchen and Leonid(2021)}]{muchen2021referring}
Muchen L, Leonid S (2021) {Referring Transformer}: A one-step approach to multi-task visual grounding. In: NeurIPS

\bibitem[{Nagaraja et~al.(2016)Nagaraja, Morariu, and Davis}]{nagaraja2016modeling}
Nagaraja VK, Morariu VI, Davis LS (2016) Modeling context between objects for referring expression understanding. In: ECCV

\bibitem[{OpenAI(2022)}]{chatgpt2022}
OpenAI (2022) Chatgpt: Optimizing language models for dialogue. \urlprefix\url{https://openai.com/blog/chatgpt}, \url{https://openai.com/blog/chatgpt}

\bibitem[{OpenAI(2023)}]{openai2023gpt4}
OpenAI (2023) {GPT-4} technical report. arXiv preprint arXiv:230308774

\bibitem[{Ouyang-Zhang et~al.(2022)Ouyang-Zhang, Cho, Zhou, and Kr{\"a}henb{\"u}hl}]{ouyangzhang2022nms}
Ouyang-Zhang J, Cho JH, Zhou X, Kr{\"a}henb{\"u}hl P (2022) {NMS} strikes back. arXiv preprint arXiv:221206137

\bibitem[{Plummer et~al.(2015)Plummer, Wang, Cervantes, Caicedo, Hockenmaier, and Lazebnik}]{plummer2015flickr30k}
Plummer BA, Wang L, Cervantes CM, Caicedo JC, Hockenmaier J, Lazebnik S (2015) {Flickr30K Entities}: Collecting region-to-phrase correspondences for richer image-to-sentence models. In: CVPR

\bibitem[{Radford et~al.(2018)Radford, Narasimhan, Salimans, Sutskever et~al.}]{radford2018improving}
Radford A, Narasimhan K, Salimans T, Sutskever I, et~al. (2018) Improving language understanding by generative pre-training. OpenAI blog

\bibitem[{Radford et~al.(2019)Radford, Wu, Child, Luan, Amodei, Sutskever et~al.}]{radford2019language}
Radford A, Wu J, Child R, Luan D, Amodei D, Sutskever I, et~al. (2019) Language models are unsupervised multitask learners. OpenAI blog

\bibitem[{Radford et~al.(2021)Radford, Kim, Hallacy, Ramesh, Goh, Agarwal, Sastry, Askell, Mishkin, Clark et~al.}]{radford2021learning}
Radford A, Kim JW, Hallacy C, Ramesh A, Goh G, Agarwal S, Sastry G, Askell A, Mishkin P, Clark J, et~al. (2021) Learning transferable visual models from natural language supervision. In: ICML

\bibitem[{Raffel et~al.(2020)Raffel, Shazeer, Roberts, Lee, Narang, Matena, Zhou, Li, and Liu}]{raffel2020exploring}
Raffel C, Shazeer N, Roberts A, Lee K, Narang S, Matena M, Zhou Y, Li W, Liu PJ (2020) Exploring the limits of transfer learning with a unified text-to-text transformer. JMLR

\bibitem[{Rasheed et~al.(2022)Rasheed, Maaz, Khattak, Khan, and Khan}]{rasheed2022bridging}
Rasheed H, Maaz M, Khattak MU, Khan S, Khan FS (2022) Bridging the gap between object and image-level representations for open-vocabulary detection. In: NeurIPS

\bibitem[{Ren et~al.(2015)Ren, He, Girshick, and Sun}]{ren2015faster}
Ren S, He K, Girshick R, Sun J (2015) {Faster R-CNN}: Towards real-time object detection with region proposal networks. In: NeurIPS

\bibitem[{Rezatofighi et~al.(2019)Rezatofighi, Tsoi, Gwak, Sadeghian, Reid, and Savarese}]{rezatofighi2019generalized}
Rezatofighi H, Tsoi N, Gwak J, Sadeghian A, Reid I, Savarese S (2019) Generalized intersection over union: A metric and a loss for bounding box regression. In: CVPR

\bibitem[{Shao et~al.(2019)Shao, Li, Zhang, Peng, Yu, Zhang, Li, and Sun}]{shao2019objects365}
Shao S, Li Z, Zhang T, Peng C, Yu G, Zhang X, Li J, Sun J (2019) {Objects365}: A large-scale, high-quality dataset for object detection. In: CVPR

\bibitem[{Shen et~al.(2023)Shen, Song, Tan, Li, Lu, and Zhuang}]{shen2023hugginggpt}
Shen Y, Song K, Tan X, Li D, Lu W, Zhuang Y (2023) {HuggingGPT}: Solving ai tasks with chatgpt and its friends in huggingface. arXiv preprint arXiv:230317580

\bibitem[{Shrivastava and Gupta(2016)}]{shrivastava2016contextual}
Shrivastava A, Gupta A (2016) Contextual priming and feedback for faster r-cnn. In: ECCV

\bibitem[{Tian et~al.(2019)Tian, Shen, Chen, and He}]{tian2019fcos}
Tian Z, Shen C, Chen H, He T (2019) {FCOS}: Fully convolutional one-stage object detection. In: CVPR

\bibitem[{Touvron et~al.(2023)Touvron, Lavril, Izacard, Martinet, Lachaux, Lacroix, Rozi{\`e}re, Goyal, Hambro, Azhar et~al.}]{touvron2023llama}
Touvron H, Lavril T, Izacard G, Martinet X, Lachaux MA, Lacroix T, Rozi{\`e}re B, Goyal N, Hambro E, Azhar F, et~al. (2023) {LLaMA}: Open and efficient foundation language models. arXiv preprint arXiv:230213971

\bibitem[{Tsimpoukelli et~al.(2021)Tsimpoukelli, Menick, Cabi, Eslami, Vinyals, and Hill}]{tsimpoukelli2021multimodal}
Tsimpoukelli M, Menick JL, Cabi S, Eslami S, Vinyals O, Hill F (2021) Multimodal few-shot learning with frozen language models. In: NeurIPS

\bibitem[{Vaswani et~al.(2017)Vaswani, Shazeer, Parmar, Uszkoreit, Jones, Gomez, Kaiser, and Polosukhin}]{vaswani2017attention}
Vaswani A, Shazeer N, Parmar N, Uszkoreit J, Jones L, Gomez AN, Kaiser {\L}, Polosukhin I (2017) Attention is all you need. In: NeurIPS

\bibitem[{Wang et~al.(2023{\natexlab{a}})Wang, Zhang, Chu, Cao, Zhou, Wu, Wang, He, and Lin}]{wang2023v3det}
Wang J, Zhang P, Chu T, Cao Y, Zhou Y, Wu T, Wang B, He C, Lin D (2023{\natexlab{a}}) {V3Det}: Vast vocabulary visual detection dataset. arXiv preprint arXiv:230403752

\bibitem[{Wang et~al.(2023{\natexlab{b}})Wang, Dai, Chen, Huang, Li, Zhu, Hu, Lu, Lu, Li et~al.}]{wang2022internimage}
Wang W, Dai J, Chen Z, Huang Z, Li Z, Zhu X, Hu X, Lu T, Lu L, Li H, et~al. (2023{\natexlab{b}}) {InternImage}: Exploring large-scale vision foundation models with deformable convolutions. In: CVPR

\bibitem[{Wang et~al.(2022)Wang, Lu, Li, Tao, Guo, Gong, and Liu}]{wang2022cris}
Wang Z, Lu Y, Li Q, Tao X, Guo Y, Gong M, Liu T (2022) {CRIS}: Clip-driven referring image segmentation. In: CVPR

\bibitem[{Wu et~al.(2023{\natexlab{a}})Wu, Yin, Qi, Wang, Tang, and Duan}]{wu2023visual}
Wu C, Yin S, Qi W, Wang X, Tang Z, Duan N (2023{\natexlab{a}}) {Visual ChatGPT}: Talking, drawing and editing with visual foundation models. arXiv preprint arXiv:230304671

\bibitem[{Wu et~al.(2023{\natexlab{b}})Wu, Li, Ding, Li, Cheng, Tong, and Loy}]{wu2023betrayed}
Wu J, Li X, Ding H, Li X, Cheng G, Tong Y, Loy CC (2023{\natexlab{b}}) Betrayed by captions: Joint caption grounding and generation for open vocabulary instance segmentation. arXiv preprint arXiv:230100805

\bibitem[{Wu et~al.(2023{\natexlab{c}})Wu, Zhang, Jin, Liu, and Loy}]{wu2023aligning}
Wu S, Zhang W, Jin S, Liu W, Loy CC (2023{\natexlab{c}}) Aligning bag of regions for open-vocabulary object detection. In: CVPR

\bibitem[{Wu et~al.(2023{\natexlab{d}})Wu, Zhu, Zhao, and Li}]{wu2023cora}
Wu X, Zhu F, Zhao R, Li H (2023{\natexlab{d}}) {CORA}: Adapting clip for open-vocabulary detection with region prompting and anchor pre-matching. In: CVPR

\bibitem[{Yang et~al.(2022)Yang, Wang, Tang, Chen, Zhao, and Torr}]{yang2022lavt}
Yang Z, Wang J, Tang Y, Chen K, Zhao H, Torr PH (2022) {LAVT}: Language-aware vision transformer for referring image segmentation. In: CVPR

\bibitem[{Yang et~al.(2023)Yang, Li, Wang, Lin, Azarnasab, Ahmed, Liu, Liu, Zeng, and Wang}]{yang2023mm}
Yang Z, Li L, Wang J, Lin K, Azarnasab E, Ahmed F, Liu Z, Liu C, Zeng M, Wang L (2023) {MM-REACT}: Prompting chatgpt for multimodal reasoning and action. arXiv preprint arXiv:230311381

\bibitem[{Young et~al.(2014)Young, Lai, Hodosh, and Hockenmaier}]{young2014image}
Young P, Lai A, Hodosh M, Hockenmaier J (2014) From image descriptions to visual denotations: New similarity metrics for semantic inference over event descriptions. TACL

\bibitem[{Yu et~al.(2016)Yu, Poirson, Yang, Berg, and Berg}]{yu2016modeling}
Yu L, Poirson P, Yang S, Berg AC, Berg TL (2016) Modeling context in referring expressions. In: ECCV

\bibitem[{Yu et~al.(2022)Yu, Iter, Wang, Xu, Ju, Sanyal, Zhu, Zeng, and Jiang}]{yu2023generate}
Yu W, Iter D, Wang S, Xu Y, Ju M, Sanyal S, Zhu C, Zeng M, Jiang M (2022) Generate rather than retrieve: Large language models are strong context generators. In: ICLR

\bibitem[{Zang et~al.(2022)Zang, Li, Zhou, Huang, and Loy}]{zang2022open}
Zang Y, Li W, Zhou K, Huang C, Loy CC (2022) Open-vocabulary detr with conditional matching. In: ECCV

\bibitem[{Zareian et~al.(2021)Zareian, Rosa, Hu, and Chang}]{zareian2021open}
Zareian A, Rosa KD, Hu DH, Chang SF (2021) Open-vocabulary object detection using captions. In: CVPR

\bibitem[{Zhang et~al.(2022{\natexlab{a}})Zhang, Zhang, Hu, Chen, Li, Dai, Wang, Yuan, Hwang, and Gao}]{zhang2022glipv2}
Zhang H, Zhang P, Hu X, Chen YC, Li LH, Dai X, Wang L, Yuan L, Hwang JN, Gao J (2022{\natexlab{a}}) {GLIPv2}: Unifying localization and vision-language understanding. In: NeurIPS

\bibitem[{Zhang et~al.(2023)Zhang, Li, Liu, Zhang, Su, Zhu, Ni, and Shum}]{zhang2022dino}
Zhang H, Li F, Liu S, Zhang L, Su H, Zhu J, Ni LM, Shum HY (2023) {DINO}: Detr with improved denoising anchor boxes for end-to-end object detection. In: ICLR

\bibitem[{Zhang et~al.(2022{\natexlab{b}})Zhang, Roller, Goyal, Artetxe, Chen, Chen, Dewan, Diab, Li, Lin et~al.}]{zhang2022opt}
Zhang S, Roller S, Goyal N, Artetxe M, Chen M, Chen S, Dewan C, Diab M, Li X, Lin XV, et~al. (2022{\natexlab{b}}) {OPT}: Open pre-trained transformer language models. arXiv preprint arXiv:220501068

\bibitem[{Zhong et~al.(2022)Zhong, Yang, Zhang, Li, Codella, Li, Zhou, Dai, Yuan, Li et~al.}]{zhong2022regionclip}
Zhong Y, Yang J, Zhang P, Li C, Codella N, Li LH, Zhou L, Dai X, Yuan L, Li Y, et~al. (2022) {RegionCLIP}: Region-based language-image pretraining. In: CVPR

\bibitem[{Zhou et~al.(2022)Zhou, Girdhar, Joulin, Kr{\"a}henb{\"u}hl, and Misra}]{zhou2022detecting}
Zhou X, Girdhar R, Joulin A, Kr{\"a}henb{\"u}hl P, Misra I (2022) Detecting twenty-thousand classes using image-level supervision. In: ECCV

\bibitem[{Zhu et~al.(2021)Zhu, Su, Lu, Li, Wang, and Dai}]{zhudeformable}
Zhu X, Su W, Lu L, Li B, Wang X, Dai J (2021) {Deformable DETR}: Deformable transformers for end-to-end object detection. In: ICLR

\end{thebibliography}

\end{document}